\pdfoutput=1
\PassOptionsToPackage{table,xcdraw}{xcolor}
\documentclass[11pt]{article}
\usepackage[]{ACL}
\usepackage{graphicx}
\usepackage{times}
\usepackage{latexsym}
\usepackage[T1]{fontenc}
\usepackage[utf8]{inputenc}
\usepackage{microtype}
\usepackage{inconsolata}
\usepackage{amsmath}
\usepackage{amssymb}
\usepackage{newunicodechar}
\newunicodechar{，}{,}
\usepackage{booktabs}
\usepackage{xcolor}
\usepackage{tabularx}
\usepackage{placeins}
\usepackage{stfloats}
\usepackage{multirow}
\usepackage{pifont}
\usepackage{hyperref}
\usepackage{mathrsfs}
\usepackage[normalem]{ulem}
\usepackage{arydshln}
\usepackage{dashrule}
\usepackage{titlesec}
\usepackage{placeins}
\usepackage{verbatim}
\usepackage{colortbl}
\usepackage{makecell}
\usepackage{caption}
\usepackage{array}
\usepackage{enumitem}
\usepackage{comment}
\usepackage{threeparttable}
\usepackage{algorithm}
\usepackage{algpseudocode}
\usepackage{seqsplit}

\usepackage[dvipsnames]{xcolor}
\definecolor{headcol}{RGB}{255,255,255}
\usepackage{tikz}
\usepackage{pgfplots}
\pgfplotsset{compat=1.18}
\usepackage{subcaption}

\definecolor{myBlue}{HTML}{27408b}
\definecolor{myOrange}{HTML}{ba4a00}
\definecolor{myRed}{HTML}{A50C36}
\definecolor{myGreen}{HTML}{178236}
\definecolor{myGray}{HTML}{B0AEAC}
\definecolor{numGreen}{HTML}{178236}
\definecolor{numGreen}{HTML}{178236}
\usepackage{fontawesome5}

\usepackage{xcolor}
\definecolor{barBlueDark}{HTML}{193CB8}
\definecolor{barBlueLight}{HTML}{2B7FFF}

\definecolor{barOrangeDark}{HTML}{CA3519}
\definecolor{barOrangeLight}{HTML}{FF692A}
\definecolor{barOrangeLighter}{HTML}{FFB86A}

\usepackage[most]{tcolorbox}

\newtcbox{\entailmentBox}{
  on line,
  colframe=green,
  colback=white,
  coltext=green!80!black,
  boxrule=1pt,
  arc=0pt,
  boxsep=1pt,
  left=1pt,
  right=1pt,
  top=0.15pt,
  bottom=0.15pt
}

\newtcbox{\contradictBox}{
  on line,
  colframe=red,
  colback=white,
  coltext=red!80!black,
  boxrule=1pt,
  arc=0pt,
  boxsep=1pt,
  left=1pt,
  right=1pt,
  top=0.15pt,
  bottom=0.15pt
}

\newtcbox{\neutralBox}{
  on line,
  colframe=gray,
  colback=white,
  coltext=gray!80!black,
  boxrule=1pt,
  arc=0pt,
  boxsep=1pt,
  left=1pt,
  right=1pt,
  top=0.15pt,
  bottom=0.15pt
}

\newtcbox{\contradicotryTextBox}{
  on line,
  colframe=red!10,
  colback=red!10,
  coltext=black,
  boxrule=0pt,
  arc=0pt,
  boxsep=0pt,
  left=1pt,
  right=1pt,
  top=0pt,
  bottom=0pt
}

\title{\textsc{eTracer}: Polarity-Aware Evidence Grounding}

\author{
\textbf{Bohao Chu}\textsuperscript{1,*} \;\;\;
\textbf{Qianli Wang}\textsuperscript{2} \;\;\;
\textbf{Hendrik Damm}\textsuperscript{3} \;\;\; \\
\textbf{Shuning Zhang}\textsuperscript{4} \;\;\;  
\textbf{Hui Wang}\textsuperscript{1} \;\;\;
\textbf{Andrea Papenmeier}\textsuperscript{5} \;\;\;
\textbf{Georg Lodde}\textsuperscript{6} \;\;\; \\ 
\textbf{Elisabeth Livingstone}\textsuperscript{6} \;\;\;
\textbf{Christoph M. Friedrich}\textsuperscript{3} \;\;\;
\textbf{Norbert Fuhr}\textsuperscript{1} \\
\textsuperscript{1}University of Duisburg-Essen,
\textsuperscript{2}Technische Universit\"at Berlin,  \\ 
\textsuperscript{3}University of Applied Sciences and Arts Dortmund,  \\
\textsuperscript{4}Tsinghua University,
\textsuperscript{5}University of Twente,
\textsuperscript{6}University Hospital Essen, \\
\texttt{\textsuperscript{*} \faEnvelope \; bohao.chu@qq.com}
}

\begin{document}
\maketitle

\begin{abstract}
Evidence grounding makes generated content verifiable by linking statements to their sources. However, effective verifiability requires two properties that existing methods fail to jointly provide: \textit{verification granularity}, the ability to localize fine-grained evidence for each statement, and \textit{verification completeness}, the ability to capture both support and contradiction. Existing methods either operate at granularities too coarse for rapid inspection or are support-only. We formalize \textit{polarity-aware evidence grounding}: linking each response sentence to its contextual evidence with a supporting, contradicting, or irrelevant label. We construct \textsc{eTracer-Bench}, a human-annotated benchmark with 7{,}905 supporting and contradicting sentence pairs across seven domains, covering both explicit and implicit contradictions. We further propose \textsc{eTracer}, a post-hoc framework whose context-aware signed scorer recovers the full evidence map in a single forward pass per response, with localization and polarity emerging jointly from one mechanism. \textsc{eTracer} achieves the best overall Citation~F1 of $82.39$, driven by a Contradict~F1 of $89.20$. A pre-registered within-subjects user study ($n{=}30$) shows that polarity-aware grounding nearly doubles contradiction detection over the document-level baseline ($27.9\%$$\to$$51.9\%$, Cohen's $d{=}0.82$, $p{=}0.0003$), and suggests a false-completeness bias under which support-only grounding fails to improve over this baseline ($23.4\%$ vs.\ $27.9\%$). Dataset and code: \url{https://github.com/chubohao/eTracer}.
\end{abstract}

\section{Introduction}

\begin{figure}[t]
    \centering
    \includegraphics[width=\linewidth]{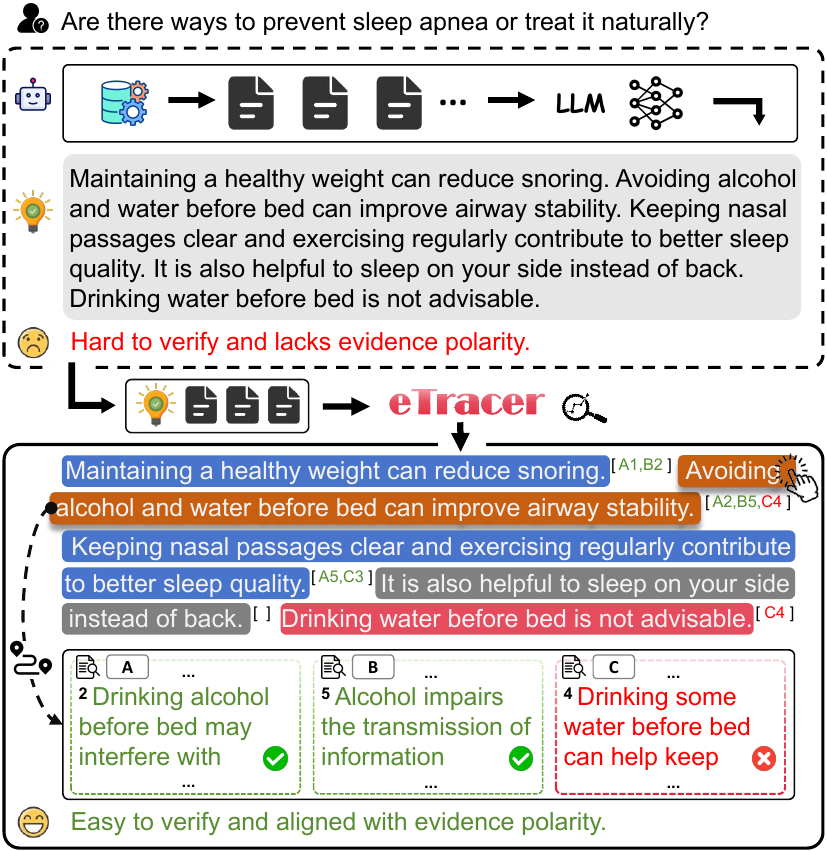}
    \vspace{-10pt}
    \caption{\textsc{eTracer} grounds each response sentence in its supporting and contradicting context sentences.}
    \label{fig:etracer}
    \vspace{-15pt}
\end{figure}

Retrieval-augmented generation (RAG) produces responses conditioned on retrieved documents, yet these responses remain difficult to verify: users cannot efficiently determine which statements are supported, contradicted, or ungrounded \citep{min-etal-2023-factscore, gao-etal-2023-enabling}. This problem is acute when retrieved documents contain conflicting or outdated information \citep{gupta2024overview}. Effective verifiability requires two properties that existing methods fail to jointly provide: \textit{verification granularity}, the ability to localize fine-grained evidence for each statement, and \textit{verification completeness}, the ability to capture both support and contradiction.

Regarding \textit{verification granularity}, existing methods are too coarse on at least one side of the grounding link. On the evidence side, both response-to-passage methods~\citep{perplexity2024, gao-etal-2023-enabling} and claim-to-passage methods~\citep{min-etal-2023-factscore, tang-etal-2024-minicheck} ground statements against entire passages, forcing users to scan long text for the relevant sentences. On the response side, claim-to-passage methods additionally decompose responses into atomic claims, a step that introduces decomposition errors and produces units misaligned with how users read \cite{wanner-etal-2024-closer, gunjal-durrett-2024-molecular}. Sentences are the natural unit of both writing and verification, offering a balance between precision and human readability. We therefore adopt \textit{sentence-to-sentence} grounding, linking each response sentence directly to the contextual sentences that evidence it.

Regarding \textit{verification completeness}, the dominant paradigms are \textit{support-only} grounding~\citep{honovich-etal-2022-true-evaluating, min-etal-2023-factscore, xie-etal-2024-doclens} and polarity-collapsed consistency scoring~\citep{laban-etal-2022-summac, zha-etal-2023-alignscore}; neither separates ungrounded statements from \textit{actively contradicted} ones. This blind spot is consequential when sources conflict, as in evidence-based medicine~\citep{sackett1996evidence} and multi-outlet news synthesis~\citep{fabbri-etal-2019-multi}. Natural language inference (NLI) models do classify polarity, but only on isolated sentence pairs~\citep{williams-etal-2018-broad}, and this pair-wise paradigm faces two issues in RAG. First, it presupposes that the relevant context sentences have already been localized. Second, scoring pairs independently discards two cross-sentence signals critical for polarity classification: the \textit{contrastive cue} between co-occurring supporting and contradicting sentences of similar surface form, and \textit{cross-sentence aggregation}, where evidence is distributed across multiple sentences that jointly establish support or imply a contradiction none states directly.

We present \textsc{eTracer}, a post-hoc framework whose context-aware signed scorer recovers the full evidence map in a single forward pass per response, with localization and polarity emerging jointly from one mechanism. As illustrated in \autoref{fig:etracer}, the resulting signed map grounds each response sentence in its supporting and contradicting context sentences, directly addressing the granularity and completeness gaps identified above. The main contributions are as follows (C1–4):

\vspace{0.05cm}
\noindent\textbf{C1:} We formalize \textit{polarity-aware evidence grounding}, a sentence-to-sentence task that makes RAG responses verifiable through fine-grained evidence alignment with their sources, jointly localizing evidence and classifying its polarity (\S\ref{sec:task}).

\vspace{0.05cm}
\noindent\textbf{C2:} We construct \textsc{eTracer-Bench}, a human-annotated dataset of 7{,}905 sentence-to-sentence evidence pairs across seven domains and two contradiction difficulties (explicit and implicit), filling a gap in existing benchmarks that lack three-way polarity at this fine-grained granularity (\S\ref{sec:dataset}).

\vspace{0.05cm}
\noindent\textbf{C3:} We propose \textsc{eTracer}, whose context-aware scorer achieves the best overall Citation~F1 of $82.39$, driven by a Contradict~F1 of $89.20$ ($+19.16$ over its matched pair-wise counterpart) (\S\ref{sec:framework}--\S\ref{sec:results}).

\vspace{0.05cm}
\noindent\textbf{C4:} A pre-registered user study ($n{=}30$) confirms our primary hypothesis that polarity-aware grounding nearly doubles contradiction identification over the document-level baseline ($d{=}0.82$), and shows signs of a \textit{false-completeness bias} under which support-only grounding offers no such gain (\S\ref{sec:user_study}).

\section{Related Work}
\label{sec:related_work}

\subsection{Evidence Granularity}
\label{sec:rw_efficiency}
Finer-grained evidence grounding enables faster human verification \citep{slobodkin-etal-2024-attribute, chu2026pcoanewbenchmarkmedical}. Existing methods vary along two orthogonal axes: the response unit being verified and the evidence unit it is grounded against. Traditional RAG systems such as Perplexity and Microsoft Copilot link response sentences to entire pages~\citep{perplexity2024, bingchat2024}. FActScore~\citep{min-etal-2023-factscore} and MiniCheck~\citep{tang-etal-2024-minicheck} decompose responses into atomic claims but verify them against entire passages, forcing users to scan the full source to confirm any claim. LongCite~\citep{zhang-etal-2025-longcite}, TracSum \cite{chu-etal-2025-tracsum}, and AGREE~\citep{slobodkin-etal-2024-attribute} achieve sentence-to-sentence grounding but remain exclusively support-oriented. \textsc{eTracer} adopts the same granularity and additionally resolves polarity, distinguishing support from contradiction.

\subsection{Polarity Classification}
\label{sec:rw_completeness}
Distinguishing whether evidence supports or contradicts a statement is essential for complete verification, especially when sources conflict. TRUE~\citep{honovich-etal-2022-true-evaluating}, DocLens~\citep{xie-etal-2024-doclens}, and LongCite~\citep{zhang-etal-2025-longcite} reduce the output to binary support labels, whether via NLI or LLM-based consistency judgments, collapsing contradiction into the unsupported category. SummaC~\citep{laban-etal-2022-summac} and AlignScore~\citep{zha-etal-2023-alignscore} can in principle detect contradiction, but aggregate pair-wise scores into a single consistency judgment, discarding sentence-level localization. \textsc{eTracer} jointly localizes evidence and resolves polarity in a single forward pass, recovering both signals existing methods leave behind.

\subsection{Evaluation Benchmarks}
\label{sec:rw_benchmarks}
Evaluating polarity-aware evidence grounding requires benchmarks that provide both fine-grained evidence annotations and three-way polarity labels. \autoref{tab:benchmark_comparison} summarizes existing RAG benchmarks along these two dimensions; none satisfies both. LFRQA~\citep{han-etal-2024-rag} is human-annotated but offers only document-level citations without contradiction labels. ALCE~\citep{gao-etal-2023-enabling} and HAGRID~\citep{kamalloo2023hagrid} ground responses at passage granularity and remain support-only. LongCite~\citep{zhang-etal-2025-longcite} reaches sentence-to-sentence granularity but still lacks contradiction labels, and its annotations are produced by GPT-4o rather than humans. \textsc{eTracer-Bench} (\S\ref{sec:dataset}) closes both gaps with human-annotated sentence-to-sentence evidence links spanning two contradiction difficulty levels (explicit and implicit).

\begin{table}[t]
\centering\small\setlength{\tabcolsep}{3pt}
\begin{tabular}{lcccc}
\toprule
\rowcolor{headcol}
\textbf{Benchmark} & \textbf{Resp.} & \textbf{Evid.} & \textbf{Polarity} & \textbf{Annot.} \\
\midrule
LFRQA                  & sent. & doc.  & S/U   & human     \\
ALCE                   & sent. & pass. & S/U   & NLI       \\
HAGRID                 & sent. & pass. & S/U   & human+LLM \\
LongCite               & sent. & sent. & S/U   & GPT-4o    \\
\midrule
\textsc{eTracer-Bench} & sent. & sent. & S/C/I & human \\
\bottomrule
\end{tabular}
\caption{Benchmark comparison. \textbf{Resp.}/\textbf{Evid.}: granularity of the response/evidence unit (sent.=sentence, doc.=document, pass.=passage). \textbf{Polarity}: S=support, C=contradict, I=irrelevant, U=unsupported. \textbf{Annot.}: annotation source.}
\label{tab:benchmark_comparison}
\vspace{-10pt}
\end{table}

\section{Task Formulation}
\label{sec:task}
We formalize \textit{polarity-aware evidence grounding} at sentence-to-sentence granularity. Given a generated response and its source context, the task is to (i)~localize the context sentences that serve as evidence for each response sentence, and (ii)~classify each link as supporting, contradicting, or irrelevant. Below we define the grounding function (\S\ref{sec:grounding_definition}).

\subsection{Polarity-Aware Evidence Grounding}
\label{sec:grounding_definition}

Let $\mathcal{R}=\{r_i\}_{i=1}^{n}$ denote a generated response segmented into $n$ sentences, and let $\mathcal{S}=\{s_j\}_{j=1}^{m}$ denote its source context segmented into $m$ sentences. Polarity-aware evidence grounding determines, for every response--context sentence pair, whether the context sentence provides supporting evidence, contradicting evidence, or no relevant evidence. Formally, a \textit{polarity-aware grounding function} $\phi:\mathcal{R}\times\mathcal{S}\rightarrow\{-1,0,+1\}$ assigns a signed label to each pair $(r_i,s_j)$ according to their evidential relationship:
\begin{equation*}
\phi(r_i,s_j)=
\begin{cases}
+1, & \text{if } s_j \text{ supports } r_i,\\
-1, & \text{if } s_j \text{ contradicts } r_i,\\
0,  & \text{if } s_j \text{ is irrelevant to } r_i.
\end{cases}
\end{equation*}

\noindent Applying $\phi$ to all $n\times m$ sentence pairs yields a matrix $\Phi\in\{-1,0,+1\}^{n\times m}$, which we call the \textit{signed evidence map}. Its rows correspond to response sentences and its columns to context sentences. Nonzero entries localize the evidence associated with each response sentence, while their signs distinguish supporting from contradicting evidence. The map therefore represents evidence localization and polarity within a single sentence-level structure.

\section{\textsc{eTracer-Bench}}
\label{sec:dataset}
As shown in \autoref{tab:benchmark_comparison}, no existing benchmark jointly provides sentence-to-sentence granularity and three-way polarity annotation. We therefore construct \textsc{eTracer-Bench}, a human-annotated benchmark of 7{,}905 sentence-to-sentence evidence pairs across seven domains with two levels of contradiction difficulty (explicit and implicit). Full construction details are provided in Appendix~\ref{sec:bench_construction_appendix}.

\vspace{0.10cm}
\noindent\textbf{Base corpus.} We extend LFRQA~\citep{han-etal-2024-rag}, a RAG-QA dataset of human-written long-form answers with document-level citations. We sample 100 responses per domain, yielding 700 responses across seven domains: biomedical (\textsc{BioASQ}), finance (\textsc{FiQA}), and the LoTTE subsets \textsc{Tech}, \textsc{Sci}, \textsc{Life}, \textsc{Writing}, and \textsc{Rec}.

\vspace{0.10cm}
\noindent\textbf{Annotation protocol.} For each response sentence, two trained annotators independently identify \textsc{Support} and \textsc{Contradict} sentences from the cited documents, with all remaining sentences treated as \textsc{Irrelevant}. A third annotator adjudicates the $7.3\%$ disagreements. Inter-annotator agreement~\citep{cohen1960coefficient} is $\kappa{=}0.87$ on evidence identification (evidence vs.\ irrelevant) and $\kappa{=}0.92$ on polarity classification among the sentences both annotators marked as evidence (Appendix~\ref{sec:natural_appendix}). This yields $4{,}565$ supporting and only $94$ naturally occurring contradicting pairs (see \autoref{tab:dataset_stats}), motivating the adversarial augmentation below.

\vspace{0.10cm}
\noindent\textbf{Dataset split.} We split the $700$ responses at the response level, stratified by domain, into train/dev/test with a $70/10/20$ per-domain ratio. Response-level splitting ensures each response and its source documents appear in exactly one split, preventing context leakage. The dev split is used exclusively for model selection, hyperparameter tuning, and few-shot demonstration sampling; all reported results are on the held-out test split.

\vspace{0.10cm}
\noindent\textbf{Contradiction construction.} To compensate for the scarcity of natural contradictions, we adversarially augment the corpus at two difficulty levels: \textbf{explicit} contradictions use surface-level alterations (negation, antonym substitution, factual replacement), while \textbf{implicit} ones introduce logically incompatible statements that preserve the topic but require multi-step inference. Each constructed paragraph is placed in a new document and appended to the context, which is then shuffled to simulate conflicting retrieval while eliminating positional bias. To prevent stylistic leakage, three disjoint annotators construct contradictions for the train/dev/test splits, with a fourth performing blind verification (Appendix~\ref{sec:contra_construction_appendix}). This yields $1{,}205$ explicit and $2{,}041$ implicit contradicting pairs (\autoref{tab:dataset_stats}).

\begin{table}[t]
\centering\small\setlength{\tabcolsep}{2.5pt}
\begin{tabular*}{\columnwidth}{@{\extracolsep{\fill}}lcccccc}
\toprule
\rowcolor{headcol}
\textbf{Domain} & \textbf{Resp.} & \textbf{Ctx} & \textbf{S} & \textbf{C$_n$} & \textbf{C$_{e}$} & \textbf{C$_{i}$} \\
\midrule
BioASQ      & 163 & 2{,}701 & 412 &  1 & 144 & 129 \\
FiQA        & 285 & 2{,}753 & 749 &  16 & 179 & 393 \\
LoTTE-Life  & 315 & 2{,}601 & 821 &  21 & 200 & 398 \\
LoTTE-Rec   & 214 & 1{,}899 & 585 &  12 & 154 & 238 \\
LoTTE-Sci   & 233 & 2{,}344 & 686 &  16 & 159 & 300 \\
LoTTE-Tech  & 266 & 2{,}174 & 634 &  12 & 173 & 327 \\
LoTTE-Writ  & 260 & 2{,}322 & 678 &  16 & 196 & 256 \\
\midrule
Total       & 1{,}736 & 16{,}794 & 4{,}565 & 94 & 1{,}205 & 2{,}041 \\
\bottomrule
\end{tabular*}
\caption{Statistics of \textsc{eTracer-Bench}. Resp.: response sentences. Ctx: context sentences. S: supporting pairs. C$_n$: naturally contradicting pairs. C$_e$/C$_i$: adversarially constructed explicit/implicit contradicting pairs.}
\label{tab:dataset_stats}
\vspace{-10pt}
\end{table}

\section{\textsc{eTracer} Framework}
\label{sec:framework}
\textsc{eTracer} rests on a structural correspondence: the task output is an $n \times m$ signed matrix over response--context sentence pairs, and a transformer encoder over the concatenated sequence already produces $n \times m$ boundary-token interactions under full self-attention. We exploit this directly. Rather than classifying each pair $(r_i, s_j)$ in isolation, \textsc{eTracer} encodes the full response and full context in a single forward pass and reads off all pairs from their boundary-token representations. This preserves two cross-sentence signals that pair-wise scoring discards by construction. The first is the \textit{contrastive cue} between co-occurring supporting and contradicting sentences of similar surface form; the second is \textit{cross-sentence aggregation} of evidence distributed across multiple sentences. Both emerge from vanilla attention within the encoded sequence (validated in \S\ref{sec:contradiction_results} and \S\ref{sec:ablation_training}), while inference collapses from $n \times m$ forward passes to one.

\subsection{Context-Aware Signed Scoring}
\label{sec:signed_scoring}
Given a response $\mathcal{R} = \{r_i\}_{i=1}^{n}$ and its context $S = \{s_j\}_{j=1}^{m}$, the context-aware scorer $\mathcal{M}_{con}$ produces a signed evidence map $\Phi \in \{+1, -1, 0\}^{n \times m}$ in a single forward pass. The response and context are packed into one input sequence with learned separators \texttt{[R\_SEP]} and \texttt{[S\_SEP]} (initialized from \texttt{[SEP]} and fine-tuned) marking sentence boundaries, and the native \texttt{[SEP]} delimiting the two blocks:
\begin{equation*}
\label{eq:input_sequence}
\resizebox{\columnwidth}{!}{$
\underbrace{\texttt{[CLS]}\; r_1 \;\texttt{[R\_SEP]}\; \cdots \; r_n \;\texttt{[R\_SEP]}}_{\text{\normalsize response}}\; \texttt{[SEP]}\; \underbrace{s_1 \;\texttt{[S\_SEP]}\; \cdots \; s_m \;\texttt{[S\_SEP]}}_{\text{\normalsize context}}
$}
\end{equation*}
\noindent Full self-attention conditions every boundary token on the entire input, so each $h_i^r$ and $h_j^s$ already encodes its sentence in the context of all others. The signed label for any pair $(r_i, s_j)$ then reduces to a two-token readout:
\begin{equation*}
p(\cdot \mid r_i, s_j) = \text{softmax}\!\left(\text{MLP}\!\left([h_i^r ; h_j^s]\right)\right),
\end{equation*}
with $\phi(r_i, s_j) = \arg\max\, p(\cdot \mid r_i, s_j) \in \{+1, -1, 0\}$ at inference. Stacking the readouts over all $n \times m$ pairs recovers the full evidence map $\Phi$ in a single forward pass: the cross-sentence reasoning that the task fundamentally demands and the cross-sentence attention that the encoder already performs are the same underlying computation.

\subsection{Training}
\label{sec:training}

\noindent\textbf{Training data.} The context-aware scorer $\mathcal{M}_{con}$ is trained on \textsc{eTracer-Bench} (train split), which provides sentence-to-sentence polarity annotations between multi-sentence responses and their full source contexts. Each training instance is a tuple $(\mathcal{R}, S, \mathbf{Y})$ where $\mathcal{R} = [r_1, \ldots, r_n]$ is the full response, $S = [s_1, \ldots, s_m]$ is the concatenation of all source document sentences, and $\mathbf{Y} \in \{+1, -1, 0\}^{n \times m}$ is the polarity label matrix, matching the inference-time input exactly. To mirror the variable document orderings returned by retrieval at deployment, we augment the train split (Appendix~\ref{sec:shuffle_augmentation}): document-order shuffling ($8$ seeds) combined with two contradiction-type views, expanding $490 \to 5{,}386$ training instances.

\vspace{0.10cm}
\noindent\textbf{Backbone and objective.} \label{sec:objective}
We initialize $\mathcal{M}_{con}$ from ModernBERT-large~\citep{warner-etal-2025-smarter} (395M parameters), whose bidirectional attention and native 8{,}192-token context window accommodate the concatenated response and context of every \textsc{eTracer-Bench} instance within a single sequence, preserving the cross-sentence signals that motivate our design (\S\ref{sec:signed_scoring}).\footnote{Longer contexts are handled by a document-level sliding-window extension, which stays within $0.8$ Citation F1 of the native-window result even at $T{=}1{,}024$ (Appendix~\ref{sec:sliding_window}).} $\mathcal{M}_{con}$ is optimized with class-weighted cross-entropy loss summed over all $n \times m$ response--context sentence pairs:

{\small
\begin{equation*}
\mathcal{L} = \sum_{i=1}^{n}\sum_{j=1}^{m} w_{y_{ij}} \, \operatorname{CE}(\hat{y}_{ij}, y_{ij})
\end{equation*}
}

\noindent where $w_{y_{ij}} \in \{w_{\text{con}}, w_{\text{irr}}, w_{\text{sup}}\}$ weighs each class according to its label $y_{ij}$; the specific values are selected on the dev split (see \S\ref{sec:experiment_setting}).

\section{Experiments}
\label{sec:setup}

We evaluate \textsc{eTracer} on the \textsc{eTracer-Bench} test split (\S\ref{sec:dataset}) and compare it against strong baselines spanning two paradigms: NLI-based pair-wise scoring and end-to-end LLM prompting. Implementation details are reported in Appendix~\ref{sec:impl_details}.

\subsection{Baselines}
\label{sec:baselines}
Our task requires three-way polarity labels at the sentence-to-sentence level (\S\ref{sec:task}), which rules out two classes of prior methods. Support-only methods~\citep{min-etal-2023-factscore, tang-etal-2024-minicheck, honovich-etal-2022-true-evaluating, zhang-etal-2025-longcite, xie-etal-2024-doclens} cannot separate contradiction from irrelevance. Aggregation-based methods~\citep{laban-etal-2022-summac, zha-etal-2023-alignscore} compute a pair-wise NLI matrix internally but collapse it into a single consistency score, discarding the sentence-level polarity our task demands; their upstream pair-wise NLI scoring is conceptually similar to our NLI baselines. The remaining applicable baselines fall into two families: pair-wise scoring and end-to-end LLM. Checkpoints and access details are in \autoref{tab:baseline_sources}.

\vspace{0.10cm}
\noindent\textbf{Pair-wise scoring.} These methods score each response-context sentence pair $(r_i, s_j)$ independently. We evaluate DeBERTa-v3-large~\citep{he2023debertav3improvingdebertausing} and BART-large~\citep{lewis-etal-2020-bart} in two configurations: off-the-shelf (fine-tuned on MNLI) and fine-tuned on the \textsc{eTracer-Bench} train split under the same class-weighted cross-entropy objective as \textsc{eTracer}. We additionally train $\mathcal{M}_{pair}$, a pair-wise variant sharing \textsc{eTracer}'s backbone, training data, and loss, and differing only in input construction, isolating cross-sentence context as the axis of variation (verified in \S\ref{sec:ablation_training}).

\vspace{0.10cm}
\noindent\textbf{LLM prompting.} As representative LLM baselines, we evaluate the closed-source GPT-4o~\citep{openai2024gpt4ocard} and the open-source Qwen3-235B-Instruct~\citep{yang2025qwen3technicalreport} under zero-shot and 7-shot settings. These two models cover both proprietary and open-source families and serve to characterize the difficulty of the task at scale, rather than to enumerate the current LLM frontier exhaustively. Each model receives the full response and full context, and outputs structured sentence-level polarity labels. Prompts are listed in Appendix~\ref{sec:prompts}.

\subsection{Metrics}
\label{sec:metrics}
We evaluate pair-level polarity labels against gold annotations. Support~F1 and Contradict~F1 score each $(r_i, s_j)$ prediction, and Citation~F1 is their macro average. We further decompose Contradict~F1 into Contradict$_e$~F1 (explicit) and Contradict$_i$~F1 (implicit) over surface-level and inference-requiring contradictions (\S\ref{sec:dataset}). Confidence intervals on the main metrics are 95\% paired bootstrap over response sentences ($10{,}000$ replicates) \cite{koehn-2004-statistical}; for the small natural-contradiction subset (Appendix~\ref{sec:natural_contra}), we additionally report Wilson 95\% CIs \cite{wilson1927probable}.

\subsection{Implementation}
\label{sec:experiment_setting}
All fine-tuned models are initialized from their respective backbones and trained on the \textsc{eTracer-Bench} train split with identical class-weighted cross-entropy loss and optimization hyperparameters; $\mathcal{M}_{con}$ and $\mathcal{M}_{pair}$ differ only in input construction (\S\ref{sec:signed_scoring}). Class weights $[w_{\text{sup}}, w_{\text{irr}}, w_{\text{con}}]$ are selected via grid search on the dev split for $\mathcal{M}_{con}$, sweeping $w_{\text{irr}} \in \{0.4, 0.5, 1, 2, 4, 6, 8, 10, 12\}$ with $w_{\text{sup}} = w_{\text{con}} = 1$; the resulting optimum $[1, 10, 1]$ up-weights the dominant irrelevant class, a counterintuitive choice reflecting that the bottleneck is reliable non-evidence rejection rather than minority-class recall (Appendix~\ref{sec:weighting_ablation}). The same value is applied uniformly to all pair-wise baselines; this choice never disadvantages any baseline by more than $0.8$ dev Citation~F1 relative to its own optimum, and in two of three cases yields improved or comparable results (Appendix~\ref{sec:weighting_baselines}). Final checkpoints are selected by dev Citation~F1.

\begin{table*}[t!]
\centering\small\setlength{\tabcolsep}{4.3pt}
\begin{tabular}{@{}llccc|ccccccc@{}}
\toprule
\textbf{Category} & \textbf{Method} & \textbf{Arch.} & \textbf{Params} & \textbf{Infer.} & \textbf{S.P} & \textbf{S.R} & \textbf{S.F1} & \textbf{C.P} & \textbf{C.R} & \textbf{C.F1} & \textbf{Cit.F1} \\
\midrule
\multirow{4}{*}{\makecell[c]{Prompt}}
 & GPT-4o (0-shot)$^\dagger$    & Dec-only & --   & 1$^\ast$ & \textbf{88.60} & 68.96 & 77.56 & \uline{91.02} & 48.64 & 63.40 & 70.48 \\
 & GPT-4o (7-shot)$^\dagger$    & Dec-only & --   & 1$^\ast$ & \uline{85.39} & 71.95 & \uline{78.10} & 87.31 & 64.96 & 74.50 & 76.30 \\
 & Qwen3-235B (0-shot)$^\dagger$ & Dec-only & 235B & 1$^\ast$ & 79.18 & 72.95 & 75.94 & \textbf{92.42} & 70.24 & 79.82 & 77.88 \\
 & Qwen3-235B (7-shot)$^\dagger$ & Dec-only & 235B & 1$^\ast$ & 78.80 & \textbf{77.49} & \textbf{78.14} & 89.51 & \uline{81.92} & \uline{85.55} & \uline{81.85} \\
\midrule
\multirow{6}{*}{\makecell[c]{Fine-tuned}}
 & DeBERTa-v3 (MNLI)$^\ddagger$          & Enc-only & 435M & 73 & 66.03 & 34.48 & 45.30 & 52.15 & 69.92 & 59.74 & 52.52 \\
 & BART-large (MNLI)$^\ddagger$          & Enc-Dec  & 406M & 73 & 69.84 & 34.15 & 45.87 & 40.47 & 68.32 & 50.83 & 48.35 \\
 & DeBERTa-v3 (\textsc{eTB})             & Enc-only & 435M & 73 & 71.51 & \uline{73.73} & 72.60 & 74.03 & 76.64 & 75.31 & 73.96 \\
 & BART-large (\textsc{eTB})             & Enc-Dec  & 406M & 73 & 74.27 & 70.40 & 72.28 & 71.75 & 77.60 & 74.56 & 73.42 \\
\cmidrule{2-12}
 & $\mathcal{M}_{pair}$ (\textsc{eTB})   & Enc-only & 395M & 73         & 74.70 & 70.07 & 72.31 & 74.11 & 66.40 & 70.04 & 71.18 \\
 & $\mathcal{M}_{con}$ (\textsc{eTracer})& Enc-only & 395M & \textbf{1} & 84.10 & 68.63 & 75.58 & 90.59 & \textbf{87.84} & \textbf{89.20} & \textbf{82.39} \\
\bottomrule
\end{tabular}
\caption{Main grounding results on the \textsc{eTracer-Bench} test set. 
\textbf{Infer.}: forward passes per response (pair-wise requires ${\approx}73$ on average; context-aware requires $1$; $1^\ast$ denotes a single prompt for LLMs). 
\textsc{eTB}: fine-tuned on \textsc{eTracer-Bench} train split. 
S./C.: support/contradict. \textbf{Bold}: best, \uline{underline}: second best. 
$\dagger$: API-based; GPT-4o parameter count is not publicly disclosed. 
$\ddagger$: off-the-shelf, no task-specific training. 
95\% bootstrap confidence intervals are reported in \autoref{tab:main_results_ci} (Appendix~\ref{sec:ci_appendix}).}
\label{tab:main_results_ci}
\label{tab:main_results}
\vspace{-10pt}
\end{table*}

\section{Results and Analysis}
\label{sec:results}

\subsection{Main Results}
\label{sec:main_grounding}
\autoref{tab:main_results} reports polarity-aware grounding performance on the \textsc{eTracer-Bench} test set, from which three findings emerge.

\vspace{0.10cm}
\noindent\textbf{1) Context-awareness drives contradiction detection.} \textsc{eTracer} reaches $89.20$ Contradict~F1 (95\% CI $[87.31, 91.16]$), $+19.16$ over the matched pair-wise variant $\mathcal{M}_{pair}$ ($70.04$, $[66.91, 72.96]$); the two intervals are disjoint by nearly $15$ points, so the gap is statistically separable and attributable to context-aware scoring alone since backbone, data, and loss are shared. By contrast, Support~F1 improves by only $+3.27$ over $\mathcal{M}_{pair}$, with overlapping intervals ($75.58$, $[73.03, 77.98]$ vs.\ $72.31$, $[69.90, 74.74]$), consistent with support being well-served by local cues while contradiction requires comparing co-occurring sentences of similar surface form (\S\ref{sec:ablation_training}). Robustness across contradiction types follows the same pattern: \textsc{eTracer}'s implicit-vs-explicit gap is the smallest among all methods ($\Delta=-9.77$, \autoref{tab:contra_level}; \S\ref{sec:contradiction_results}).

\vspace{0.10cm}
\noindent\textbf{2) Competitive performance at two orders of magnitude fewer parameters.} \textsc{eTracer} matches Qwen3-235B 7-shot on Citation~F1 ($82.39$, $[80.78, 83.89]$ vs.\ $81.85$, $[80.31, 83.31]$; overlapping CIs, statistically on par) in a single forward pass without few-shot demonstrations, at roughly $1/600$ the parameter count. This indicates that polarity-aware grounding is amenable to specialized small architectures rather than requiring general-purpose LLM scale.

\vspace{0.10cm}
\noindent\textbf{3) Targeted supervision dominates prompt scaling.} Fine-tuning on \textsc{eTracer-Bench} lifts MNLI checkpoints by $+21.44$ (DeBERTa-v3) and $+25.07$ (BART-large) Citation~F1, dwarfing the gains from few-shot demonstrations over zero-shot ($+5.82$ for GPT-4o, $+3.97$ for Qwen3-235B). \textsc{eTracer} ($395$M, fine-tuned) remains statistically on par with Qwen3-235B 7-shot ($82.39$ vs.\ $81.85$ Citation~F1, overlapping CIs) despite a ${\sim}600{\times}$ parameter gap, indicating that for polarity-aware grounding, task-specific supervision is at least as parameter-efficient as in-context scaling.

\vspace{0.10cm}
\noindent\textbf{4) The context-aware advantage persists across training seeds.} We additionally train \textsc{eTracer} and the matched $\mathcal{M}_{pair}$ baseline with seeds $52$ and $62$, complementing the original seed $42$, using the same data-processing, optimization, and model-selection protocol. As shown in \autoref{tab:multiseed_results}, the mean Citation~F1 gain is $+10.31$ points, driven primarily by a $+17.69$ Contradict~F1 gain, while the Support~F1 gain is $+2.93$. The gain remains positive for every tested seed, indicating that the main context-aware advantage is not specific to the original seed.

\begin{table}[t]
\centering
\small
\begin{tabular*}{\columnwidth}{@{\extracolsep{\fill}}lccc@{}}
\toprule
Model & S.F1 & C.F1 & Cit.F1 \\
\midrule
\textsc{eTracer} & $75.47{\pm}0.10$ & $88.71{\pm}1.10$ & $82.09{\pm}0.58$ \\
$\mathcal{M}_{pair}$ & $72.54{\pm}0.61$ & $71.03{\pm}1.07$ & $71.79{\pm}0.81$ \\
$\Delta$ & $+2.93{\pm}0.63$ & $+17.69{\pm}1.33$ & $+10.31{\pm}0.79$ \\
\bottomrule
\end{tabular*}
\caption{Mean ${\pm}$ standard deviation across training seeds $\{42,52,62\}$. Each $\Delta$ is computed between \textsc{eTracer} and $\mathcal{M}_{pair}$ for the same seed.}
\label{tab:multiseed_results}
\vspace{-10pt}
\end{table}

\begin{table}[b!]
\vspace{-10pt}
\centering\small\setlength{\tabcolsep}{5pt}
\begin{tabular}{lcccc}
\toprule
\textbf{Method} & \textbf{C.F1} & \textbf{C$_e$.F1} & \textbf{C$_i$.F1} & $\boldsymbol{\Delta}$ \\
\midrule
GPT-4o (7-shot)                           & 74.50 & 89.50 & 61.54 & $-$27.96 \\
Qwen3-235B (7-shot)                        & \uline{85.55} & \uline{91.31} & \uline{81.48} & \uline{$-$9.83} \\
\midrule
DeBERTa-v3 (MNLI)                         & 59.74 & 67.26 & 53.30 & $-$13.96 \\
BART-large (MNLI)                         & 50.83 & 57.36 & 45.16 & $-$12.20 \\
DeBERTa-v3 (\textsc{eTB})                 & 75.31 & 83.71 & 69.35 & $-$14.36 \\
BART-large (\textsc{eTB})                 & 74.56 & 83.83 & 68.14 & $-$15.69 \\
$\mathcal{M}_{pair}$ (\textsc{eTB})       & 70.04 & 82.82 & 59.91 & $-$22.91 \\
\midrule
$\mathcal{M}_{con}$ (\textsc{eTracer})    & \textbf{89.20} & \textbf{95.18} & \textbf{85.41} & \textbf{$-$9.77} \\
\bottomrule
\end{tabular}
\caption{Contradict~F1 by contradiction type. C$_e$/C$_i$: explicit/implicit; $\Delta$: C$_i$$-$C$_e$ (smaller $=$ more robust). \textbf{Bold}/\uline{underline}: best/second best.}
\label{tab:contra_level}
\vspace{-10pt}
\end{table}

\subsection{Contradiction Detection by Type}
\label{sec:contradiction_results}

\autoref{tab:contra_level} stratifies Contradict~F1 by explicit and implicit contradictions (\S\ref{sec:dataset}). All methods handle explicit contradictions reasonably, where surface cues such as negation and antonyms suffice. Implicit contradictions widen the gap: fine-tuned pair-wise methods drop $14.4 \ldots 22.9$ points ($\Delta$ column), whereas \textsc{eTracer} drops only $9.77$ and retains $85.41$ C$_i$~F1, the smallest gap among all methods. Even Qwen3-235B 7-shot, the strongest LLM baseline ($\Delta\!=\!-9.83$), trails \textsc{eTracer} on both types ($+3.87$ C$_e$, $+3.93$ C$_i$). This robustness is the more diagnostic signal, as it cannot stem from training-set style transfer: the train/dev/test contradictions were constructed by three disjoint annotators (\S\ref{sec:dataset}). It is corroborated by two independent analyses, attention masking (\S\ref{sec:ablation_training}) and natural-vs-constructed generalization (\S\ref{sec:natural_contradictions}), confirming that context-aware scoring is the architectural ingredient closing the C$_e$/C$_i$ gap.

\subsection{Generalization to Naturally Occurring Contradictions}
\label{sec:natural_contradictions}

Only $24$ naturally occurring contradiction pairs appear in the test split, limiting the precision of real-world robustness estimates. We therefore conduct two analyses to assess whether the observed advantage extends beyond constructed contradictions.

\vspace{0.10cm}
\noindent\textbf{Generalization by contradiction structure.} The $24$ natural test pairs comprise $12$ self-contained contradictions and $12$ cross-sentence contradictions requiring additional context. As shown in \autoref{tab:natural_by_structure}, \textsc{eTracer} correctly identifies $18/24$ pairs, compared with $1/24$ for $\mathcal{M}_{pair}$; an exact paired McNemar test confirms a significant overall difference ($p{=}1.53{\times}10^{-5}$). The larger cross-sentence gap ($10/12$ vs.\ $0/12$) is consistent with context-aware evidence aggregation, although the subgroup results remain descriptive given their small size.

\begin{table}[h]
\centering
\small
\begin{tabular*}{\columnwidth}{@{\extracolsep{\fill}}lccc@{}}
\toprule
Model & All & Self & Cross-sent. \\
\midrule
\textsc{eTracer} & $18/24$ & $8/12$ & $10/12$ \\
$\mathcal{M}_{pair}$ & $1/24$ & $1/12$ & $0/12$ \\
\bottomrule
\end{tabular*}
\caption{Accuracy on natural contradictions. Self: self-contained; Cross-sent.: cross-sentence.}
\vspace{-10pt}
\label{tab:natural_by_structure}
\end{table}

\vspace{0.10cm}
\noindent\textbf{Transfer without natural-contradiction supervision.} We train an additional \textsc{eTracer} model while masking all natural contradiction pairs from the training loss and excluding them from the development-set model-selection metric, retaining all constructed contradictions and other supervision. As shown in \autoref{tab:natural_supervision}, removing natural-contradiction supervision leaves aggregate natural-test accuracy unchanged at $18/24$. This provides preliminary evidence that supervision from constructed contradictions transfers to natural conflicts, although the small test set and single training run preclude a stronger conclusion.

\begin{table}[h]
\centering
\small
\begin{tabular*}{\columnwidth}{@{\extracolsep{\fill}}lc@{}}
\toprule
Training setting & Natural accuracy \\
\midrule
Full model & $18/24$ \\
Constructed-only & $18/24$ \\
\bottomrule
\end{tabular*}
\caption{Natural-test accuracy by training supervision.}
\label{tab:natural_supervision}
\end{table}

\subsection{Domain-wise Analysis}
\label{sec:cross_domain}

\autoref{fig:domain_f1} reports per-domain Citation~F1 on the \textsc{eTracer-Bench} test split. \textsc{eTracer} outperforms $\mathcal{M}_{pair}$ in all seven domains, with Citation~F1 ranging from $79.5$ (FiQA) to $86.2$ (Lifestyle/Recreation). Gains are largest in Writing ($+15.8$), Lifestyle ($+15.6$), and Science ($+13.2$), where contradictions hinge on specific claims, mechanisms, or factual details that reward cross-sentence comparison, and smallest in BioASQ ($+6.2$), where $\mathcal{M}_{pair}$ already attains its strongest pair-wise Citation~F1 ($78.2$) and leaves less headroom. The asymmetry is sharper on contradiction alone: \textsc{eTracer} exceeds $85$ Contradict~F1 in every domain, peaking at $95.3$ (Recreation) and $93.5$ (BioASQ), whereas $\mathcal{M}_{pair}$ falls as low as $63.4$ (Writing) and $63.7$ (Science), yielding per-domain contradiction gaps up to $+25.7$ (Lifestyle) and $+24.1$ (Science) that underscore the value of context-aware scoring in analytically intensive domains.

\begin{figure}[t]
\centering
\begin{tikzpicture}
\begin{axis}[
    ybar=0.1pt,
    bar width=10pt,
    width=1.05\columnwidth,
    height=4cm,
    ymin=60, ymax=110,
    ylabel={Citation F1 (\%)},
    ylabel style={font=\small},
    symbolic x coords={Bio, Fin, Life, Rec, Sci, Tech, Writ},
    xtick=data,
    xticklabel style={font=\small},
    yticklabel style={font=\small},
    tick style={black},
    ytick={60,70,80,90},
    ymajorgrids=true,
    xmajorgrids=false,
    grid style={dotted, gray!50},
    xtick pos=left,
    ytick pos=left,
    legend style={
        at={(0.5,1.02)},
        anchor=south,
        legend columns=2,
        font=\small,
        column sep=1em,
        draw=none,
        fill=none,
    },
    enlarge x limits=0.1,
    nodes near coords,
    nodes near coords style={
        font=\scriptsize,
        rotate=90,
        anchor=west,
        yshift=0pt,
        /pgf/number format/.cd,
        fixed,
        fixed zerofill,
        precision=1
    },
]
\addplot[fill=barBlueDark, draw=white] coordinates {
    (Bio,84.4)(Fin,79.5)(Life,86.2)(Rec,86.2)(Sci,81.1)(Tech,80.5)(Writ,80.2)
};
\addlegendentry{$\mathcal{M}_{con}$ (\textsc{eTracer})}
\addplot[fill=barBlueLight, draw=white] coordinates {
    (Bio,78.2)(Fin,71.5)(Life,70.6)(Rec,74.7)(Sci,67.9)(Tech,71.7)(Writ,64.4)
};
\addlegendentry{$\mathcal{M}_{pair}$}
\end{axis}
\end{tikzpicture}
\caption{Per-domain Citation~F1 for \textsc{eTracer} and the controlled pair-wise baseline $\mathcal{M}_{pair}$. Full breakdown in Appendix~\ref{sec:domain_breakdown}.}
\label{fig:domain_f1}
\vspace{-10pt}
\end{figure}

\subsection{Ablation Study}
\label{sec:ablation_training}

\autoref{tab:ablation} isolates the contribution of context-aware scoring by selectively masking cross-sentence attention at test time. All ablations share the same trained \textsc{eTracer} model, identical input sequences, and identical decoding procedure; only the attention mask differs, so any performance change reflects the masked information flow alone.

\vspace{0.10cm}
\noindent\textbf{Masking context-side cross-attention.} When each context sentence can see the full response but not any other context sentence, Support~F1 drops by $18.11$ and Contradict~F1 by $13.62$ (Cit.F1: $-15.86$). The decline on Support reveals what we term \textit{joint evidence aggregation}: multiple context sentences each supply partial information that only jointly constitutes sufficient evidence, so disabling cross-context attention forces the model into pair-wise reasoning and erases this signal. The drop on Contradict confirms a complementary role: polarity discrimination relies on comparing co-occurring context sentences, since a supporting and a contradicting sentence often share surface form and differ only in polarity cues; without seeing both, the model loses the contrastive signal needed to tell them apart. This is the most impactful single-axis ablation, establishing context-side cross-attention as the dominant source of \textsc{eTracer}'s advantage over pair-wise scoring.

\vspace{0.10cm}
\noindent\textbf{Masking response-side cross-attention.} When each response sentence can see all context sentences but not any other response sentence, Support~F1 drops by $1.80$ and Contradict~F1 by $3.98$ (Cit.F1: $-2.89$). The modest decline suggests that response-side cross-attention contributes secondarily, likely by resolving coreference across response sentences, for instance linking a pronoun in $r_2$ back to an entity introduced in $r_1$, rather than driving polarity classification directly.

\vspace{0.10cm}
\noindent\textbf{Masking both.} Isolating context and response sentences simultaneously drops Cit.F1 by $18.59$, only marginally beyond the context-side ablation alone ($-15.86$). Since the response-side ablation costs $-2.89$ Cit.F1 in isolation and contributes $-2.73$ on top of the context-side ablation, the two signals operate near-additively: their effects accumulate without meaningful overlap, with the context side clearly dominant ($\sim\!85\%$ of the joint degradation).

\begin{table}[t]
\centering\small\setlength{\tabcolsep}{2pt}
\begin{tabular}{lcccc}
\toprule
\rowcolor{headcol}
\textbf{System} & \textbf{S.F1} & \textbf{C.F1} & \textbf{Cit.F1} & $\boldsymbol{\Delta}$ \\
\midrule
\textsc{eTracer} (full)            & \textbf{75.58} & \textbf{89.20} & \textbf{82.39} & ---     \\
\quad Mask context cross-attn.     & 57.47          & 75.58          & 66.53          & $-15.86$ \\
\quad Mask response cross-attn.    & 73.78          & 85.22          & 79.50          & $-2.89$  \\
\quad Mask both                    & 60.04          & 67.56          & 63.80          & $-18.59$ \\
\bottomrule
\end{tabular}
\caption{Ablation on the \textsc{eTracer-Bench} test set. All rows share the same trained \textsc{eTracer} model with selective attention masking at test time. $\Delta$: change in Citation F1 vs.\ full system.}
\label{tab:ablation}
\vspace{-5pt}
\end{table}

\begin{table}[t]
\centering\small\setlength{\tabcolsep}{3pt}
\begin{tabular}{lccc}
\toprule
\rowcolor{headcol}
\textbf{Condition} & \textbf{S.F1} & \textbf{C.F1} & \textbf{Cit.F1} \\
\midrule
$\mathcal{M}_{con}$ (original)   & $75.58$ & $89.20$ & $82.39$ \\
$\mathcal{M}_{con}$ (shuffled)   & $70.42${\scriptsize$\pm0.12$} & $70.35${\scriptsize$\pm1.60$} & $70.39${\scriptsize$\pm0.85$} \\
$\Delta$                         & $-5.16$ & $-18.85$ & $-12.00$ \\
\midrule
$\mathcal{M}_{pair}$ (invariant) & $72.31$ & $70.04$ & $71.18$ \\
\bottomrule
\end{tabular}
\caption{Context shuffling on the test split. Shuffled results: mean $\pm$ std over 3 seeds $\in\{42,52,62\}$. $\mathcal{M}_{pair}$ serves as a null control, invariant to shuffling by construction. Shuffled $\mathcal{M}_{con}$ converges to $\mathcal{M}_{pair}$'s native level across all pair-level metrics.}
\label{tab:shuffle}
\vspace{-10pt}
\end{table}

\vspace{0.10cm}
\noindent\textbf{Context shuffling.} To verify that \textsc{eTracer} exploits cross-sentence \textit{structure} rather than mere co-occurrence, we shuffle the order of context sentences at inference time. $\mathcal{M}_{pair}$ is invariant to this perturbation by construction and serves as a null control. Shuffling drops Contradict~F1 by $18.85$ points and Support~F1 by only $5.16$ (3 seeds $\in\{42,52,62\}$, std $\leq 1.60$; Appendix~\ref{sec:shuffling_per_seed}). The asymmetry is diagnostic: contradiction detection hinges on relative positions among co-occurring sentences, whereas support detection relies more on local sentence-pair semantics. Strikingly, shuffled $\mathcal{M}_{con}$ converges almost exactly to $\mathcal{M}_{pair}$'s native level (S.F1: $70.42$ vs.\ $72.31$; C.F1: $70.35$ vs.\ $70.04$; Cit.F1: $70.39$ vs.\ $71.18$): stripping cross-sentence structure leaves $\mathcal{M}_{con}$ at essentially the same operating point as a model structurally incapable of seeing it. The context-aware gain (\autoref{tab:main_results}) is thus \textit{entirely} mediated by cross-sentence structure, not by incidental factors such as parameter count or effective training data.

\section{User Study}
\label{sec:user_study}

Automatic metrics measure label agreement; their value rests on whether they translate to better human verification. To isolate the value of the grounding paradigm from model accuracy, all evidence highlights in the user study are drawn from gold annotations rather than from \textsc{eTracer} predictions. This prevents prediction errors from confounding the comparison and allows us to evaluate the support-only and polarity-aware paradigms under correct evidence grounding. We test this with a pre-registered\footnote{Pre-reg.: \url{https://aspredicted.org/vv3v2k.pdf}.}, within-subjects experiment ($n{=}30$, recruited via Prolific). Full protocol is described in Appendix~\ref{sec:user_study_details}.

\vspace{0.10cm}
\noindent\textbf{Design.} We compare three conditions: \textbf{(D)} \textit{document-level}, the response beside its full source without citation; \textbf{(S)} \textit{support-only}, where clicking a response sentence highlights its supporting context sentences in green; and \textbf{(E)} \textit{polarity-aware}, additionally highlighting contradicting sentences in red. Highlights in both $S$ and $E$ are drawn from gold annotations rather than from \textsc{eTracer} predictions, isolating each paradigm from model quality and preventing prediction errors from obscuring the comparison. The 15 tasks are rotated across conditions via a 6-sequence Williams Latin square, yielding 10 participants per (group, condition) cell.

\begin{figure}[t]
\centering
\begin{tikzpicture}
\begin{axis}[
    ybar=0.5pt, bar width=15pt,
    width=1.08\columnwidth, height=3.5cm,
    ymin=0, ymax=110,
    symbolic x coords={Acc.\,gold-A$\uparrow$, Acc.\,gold-B$\uparrow$, Asym.\,score$\downarrow$},
    xtick=data,
    xticklabel style={font=\scriptsize},
    yticklabel style={font=\scriptsize},
    ytick={0,25,50,75,100},
    ymajorgrids=true, grid style={dotted, gray!50},
    enlarge x limits={abs=28pt},
    axis x line*=box, axis y line*=box,
    xtick pos=left, ytick pos=left,
    legend style={at={(0.5,1.02)}, anchor=south, legend columns=3,
                  font=\scriptsize, column sep=0.4em, draw=none},
    nodes near coords, nodes near coords style={font=\tiny, anchor=south},
]
\addplot[fill=barOrangeDark, draw=white] coordinates {
    (Acc.\,gold-A$\uparrow$,73.0) (Acc.\,gold-B$\uparrow$,27.9) (Asym.\,score$\downarrow$,44.2)};
\addlegendentry{D}
\addplot[fill=barOrangeLight, draw=white] coordinates {
    (Acc.\,gold-A$\uparrow$,77.6) (Acc.\,gold-B$\uparrow$,23.4) (Asym.\,score$\downarrow$,56.3)};
\addlegendentry{S}
\addplot[fill=barOrangeLighter, draw=white] coordinates {
    (Acc.\,gold-A$\uparrow$,85.8) (Acc.\,gold-B$\uparrow$,51.9) (Asym.\,score$\downarrow$,36.3)};
\addlegendentry{E}
\end{axis}
\end{tikzpicture}
\caption{User-study results ($n{=}30$, within-subjects). S's collapsed gold-B accuracy paired with elevated gold-A accuracy shows signs of a false-completeness bias, which E eliminates while raising both accuracies.}
\label{fig:user_study}
\vspace{-15pt}
\end{figure}

\vspace{0.10cm}
\noindent\textbf{Task and Measures.} Each task pairs a RAG response with its full cited context. Participants evaluate the response sentences against the source evidence rather than assessing \textsc{eTracer} predictions. They click each response sentence, view the condition-appropriate evidence, and record a binary citation decision: \textit{(A) include as-is} or \textit{(B) do not include}. Gold-A sentences have only supporting evidence; gold-B sentences also contain contradicting evidence, where citing as-is would ignore the conflict. The 15 tasks comprise 35 sentences (15 gold-A, 20 gold-B), yielding $1{,}050$ decisions across 30 participants and 3 conditions. We report \textit{decision accuracy} by gold label and the \textit{within-response asymmetry score} (gold-A minus gold-B accuracy): scores near $0$ indicate symmetric judgment; large positive scores indicate accepting support at face value while missing conflict. The pre-registered primary contrast is $S$ vs.\ $E$ on gold-B accuracy.

\vspace{0.10cm}
\noindent\textbf{Findings.} Two patterns emerge (\autoref{fig:user_study}; full statistics in Appendix~\ref{sec:us_stats}). \textit{(F1) Polarity-aware grounding eliminates the bias and raises contradiction accuracy:} confirming our pre-registered primary hypothesis, $E$ more than doubles gold-B accuracy over $S$ ($51.9\%$ vs.\ $23.4\%$, $+28.5$ pp, $d{=}0.82$, $p{=}.0003$) and lowers the asymmetry score to $36.3$; gold-A accuracy is also higher under $E$ ($85.8\%$). The $S{\to}E$ gap meets our pre-registered threshold ($d \geq 0.8$) and survives Holm--Bonferroni correction (Holm-corrected $p{=}.0029$). \textit{(F2) Support-only grounding shows signs of a false-completeness bias:} $S$ yields numerically lower gold-B accuracy than $D$ ($23.4\%$ vs.\ $27.9\%$), though the effect is small ($d{=}0.16$, $p{=}0.135$) and our sample is underpowered to confirm it; support-only highlights may signal sufficiency, leading participants to overlook unhighlighted conflicts, and the asymmetry score under $S$ ($56.3$) exceeds that under $D$ ($44.2$).

\section{Conclusion}
\label{sec:conclusion}
We presented \textsc{eTracer}, a post-hoc framework for polarity-aware evidence grounding that jointly localizes sentence-to-sentence evidence and classifies its polarity in a single forward pass. Its context-aware signed scorer exploits a transformer encoder's self-attention, reading the full $n{\times}m$ signed evidence map off the boundary-token interactions the encoder already computes. This single-pass design surpasses a matched pair-wise baseline by $+19.16$ Contradict~F1 while matching strong LLM baselines at roughly $1/600$ their scale. A pre-registered user study shows that polarity-aware grounding significantly improves contradiction identification (Cohen's $d{=}0.82$), and further suggests a \textit{false-completeness bias} under which support-only grounding fails to improve over the no-citation baseline. We release \textsc{eTracer-Bench}, with continued maintenance, to support future work on grounding that is both fine-grained and polarity-aware.

\clearpage

\section*{Limitations}
\label{sec:limitations}

\noindent\textbf{Context length.}
\textsc{eTracer} is built on ModernBERT-large, whose native 8{,}192-token window accommodates every \textsc{eTracer-Bench} instance without truncation. Since individual retrieved documents rarely exceed this budget, the limit is reached only when the \emph{number} of retrieved documents grows large. For such cases, \textsc{eTracer} admits a natural document-level sliding-window extension: partition the documents into disjoint groups that each fit the window, concatenate every group with the full response as an independent input, and merge the resulting signed maps along the context axis into a complete $n \times m$ map. Partitioning at document boundaries preserves each document's semantic integrity and keeps context-aware scoring intact within each window, at the cost of cross-window dependencies. We validate this extension in Appendix~\ref{sec:sliding_window}: at $T{=}2{,}048$ only $3.6\%$ of test instances require multiple windows, with pooled Citation F1 unchanged from the native window (within run-to-run noise); even at the aggressive $T{=}1{,}024$ ($25.0\%$ triggered), pooled Citation F1 stays within $0.8$ points of the native result. A systematic study of long-context strategies (e.g., hierarchical scoring or retrieval-augmented chunking) is left to future work.

\vspace{0.10cm}
\noindent\textbf{Naturally occurring contradictions.}
\textsc{eTracer-Bench} contains $94$ naturally occurring sentence-to-sentence contradicting pairs, split 59/11/24 across train/dev/test. The test subset suffices for a statistically separable comparison among methods (\S\ref{sec:contradiction_results}, Appendix~\ref{sec:natural_contra}), but it is modest in size and skews toward single-sentence conflicts; the multi-document and cross-sentence cases that most directly probe context-aware reasoning are underrepresented---though the architectural advantage we report rests on the controlled $M_\text{pair}$ comparison (\S\ref{sec:ablation_training}), independent of this subset's size. Two directions follow: scaling natural-contradiction annotation by systematically retrieving genuinely conflicting sources in domains such as evidence-based medicine, with deliberate emphasis on cross-sentence and multi-document conflicts; and extending the schema to pragmatic contradiction types such as deprecation, supersession, and evidence-strength divergence.

\vspace{0.10cm}
\noindent\textbf{User study scale.}
Our within-subjects study ($n{=}30$) is powered to detect large effects (Cohen's $d \geq 0.8$) at $\alpha{=}0.05$ after Holm--Bonferroni correction, matching the primary $S{\to}E$ contrast in \S\ref{sec:user_study}. Smaller effects, such as the $D$ vs.\ $S$ difference suggestive of a false-completeness bias ($d{=}0.16$, $p{=}0.135$) or domain- and task-type-specific differences between conditions, would require substantially larger samples to confirm and are left to follow-up work.

\vspace{0.10cm}
\noindent\textbf{Training variability.} The main result tables report the run with seed $42$ for consistency with the original evaluation. We additionally train \textsc{eTracer} and $\mathcal{M}_{pair}$ with seeds $52$ and $62$; the context-aware gain remains positive across all three seeds (\autoref{tab:multiseed_results}). Although this analysis shows that the main result is not specific to seed $42$, three runs provide only a limited estimate of training variability.

\section*{Ethics Statement}
\noindent\textbf{Annotation.} Our work re-annotates \textsc{LFRQA}, a publicly released corpus distributed under the Apache License 2.0. The \textsc{BioASQ} subset is derived from publicly available biomedical literature rather than patient records and contains no personally identifiable information. Dataset annotation was conducted by an in-house team compensated at the authors' institution's standard hourly rate for student research assistants (\S\ref{sec:annotator_appendix}).

\vspace{0.1cm}
\noindent\textbf{User study.} The user study recruited participants through Prolific. Participants provided informed consent, could withdraw at any time without penalty, and received \pounds10.00 for a median completion time of approximately 38 minutes (\pounds15.85 per hour). We collected only task responses and completion times; no names, contact details, or other directly identifying information were collected. Study records were pseudonymized using Prolific identifiers, with recruitment and payment handled by the platform. Under our institution's policy, this minimal-risk study involving judgments of publicly derived text did not require formal ethics or IRB review.

\section*{Acknowledgments}
This work was supported by WisPerMed (AI for Personalized Medicine), funded by the German Research Foundation (DFG) under Research Training Group 2535. Generative AI tools were used for language editing, proofreading, and checklist preparation. All scientific claims, analyses, citations, and final text were reviewed and verified by the authors.

\bibliography{anthology-1, anthology-2, custom}

\clearpage
\onecolumn
\section*{Appendix}

\begin{itemize}[left=-5pt,label={}]
    \item \textbf{\hyperref[sec:bench_construction_appendix]{A \hspace{0.5em} Dataset Construction Details}} \dotfill \pageref*{sec:bench_construction_appendix}
    \begin{itemize}[label={}]
        \item \hyperref[sec:base_corpus_appendix]{A.1 \hspace{0.5em} Base Corpus} \dotfill \pageref*{sec:base_corpus_appendix}
        \item \hyperref[sec:annotator_appendix]{A.2 \hspace{0.5em} Annotation Team} \dotfill \pageref*{sec:annotator_appendix}
        \item \hyperref[sec:biomed_appendix]{A.3 \hspace{0.5em} Biomedical Domain Support} \dotfill \pageref*{sec:biomed_appendix}
        \item \hyperref[sec:natural_appendix]{A.4 \hspace{0.5em} Natural Annotation Stage} \dotfill \pageref*{sec:natural_appendix}
        \item \hyperref[sec:dataset_split_appendix]{A.5 \hspace{0.5em} Dataset Split} \dotfill \pageref*{sec:dataset_split_appendix}
        \item \hyperref[sec:contra_construction_appendix]{A.6 \hspace{0.5em} Adversarial Construction Stage} \dotfill \pageref*{sec:contra_construction_appendix}
        \item \hyperref[sec:corpus_stats_appendix]{A.7 \hspace{0.5em} Dataset Statistics} \dotfill \pageref*{sec:corpus_stats_appendix}
    \end{itemize}

    \item \textbf{\hyperref[sec:impl_details]{B \hspace{0.5em} Implementation Details}} \dotfill \pageref*{sec:impl_details}
    \begin{itemize}[label={}]
        \item \hyperref[sec:data_pipeline]{B.1 \hspace{0.5em} Shared Data Pipeline} \dotfill \pageref*{sec:data_pipeline}
        \item \hyperref[sec:input_construction]{B.2 \hspace{0.5em} Input Construction} \dotfill \pageref*{sec:input_construction}
        \item \hyperref[sec:training_config]{B.3 \hspace{0.5em} Optimization} \dotfill \pageref*{sec:training_config}
        \item \hyperref[sec:shuffle_augmentation]{B.4 \hspace{0.5em} Document-Order Data Augmentation} \dotfill \pageref*{sec:shuffle_augmentation}
        \item \hyperref[sec:reproducibility_appendix]{B.5 \hspace{0.5em} Reproducibility} \dotfill \pageref*{sec:reproducibility_appendix}
        \item \hyperref[sec:baseline_sources]{B.6 \hspace{0.5em} Off-the-Shelf Baselines} \dotfill \pageref*{sec:baseline_sources}
        \item \hyperref[sec:latency_protocol]{B.7 \hspace{0.5em} Inference Cost} \dotfill \pageref*{sec:latency_protocol}
    \end{itemize}

    \item \textbf{\hyperref[sec:supplementary]{C \hspace{0.5em} Supplementary Experiments}} \dotfill \pageref*{sec:supplementary}
    \begin{itemize}[label={}]
        \item \hyperref[sec:ci_appendix]{C.1 \hspace{0.5em} Main Results with Confidence Intervals} \dotfill \pageref*{sec:ci_appendix}
        \item \hyperref[sec:domain_breakdown]{C.2 \hspace{0.5em} Per-Domain Grounding Results} \dotfill \pageref*{sec:domain_breakdown}
        \item \hyperref[sec:weighting_ablation]{C.3 \hspace{0.5em} Class-Weighting for Context-Aware} \dotfill \pageref*{sec:weighting_ablation}
        \item \hyperref[sec:weighting_baselines]{C.4 \hspace{0.5em} Class-Weighting for Pair-wise Baselines} \dotfill \pageref*{sec:weighting_baselines}
        \item \hyperref[sec:shuffling_per_seed]{C.5 \hspace{0.5em} Context Shuffling: Per-Seed Results} \dotfill \pageref*{sec:shuffling_per_seed}
        \item \hyperref[sec:natural_contra]{C.6 \hspace{0.5em} Analysis on Naturally Occurring Contradictions} \dotfill \pageref*{sec:natural_contra}
        \item \hyperref[sec:aug_ablation]{C.7 \hspace{0.5em} Document-Order Augmentation Ablation} \dotfill \pageref*{sec:aug_ablation}
        \item \hyperref[sec:sliding_window]{C.8 \hspace{0.5em} Context-Window Scaling via Sliding Windows} \dotfill \pageref*{sec:sliding_window}
    \end{itemize}

    \item \textbf{\hyperref[sec:user_study_details]{D \hspace{0.5em} User Study Details}} \dotfill \pageref*{sec:user_study_details}
    \begin{itemize}[label={}]
        \item \hyperref[sec:us_protocol]{D.1 \hspace{0.5em} Protocol and Participants} \dotfill \pageref*{sec:us_protocol}
        \item \hyperref[sec:us_materials]{D.2 \hspace{0.5em} Task Materials} \dotfill \pageref*{sec:us_materials}
        \item \hyperref[sec:us_measures]{D.3 \hspace{0.5em} Measures} \dotfill \pageref*{sec:us_measures}
        \item \hyperref[sec:us_stats]{D.4 \hspace{0.5em} Full Statistical Results} \dotfill \pageref*{sec:us_stats}
    \end{itemize}

    \item \textbf{\hyperref[sec:prompts]{E \hspace{0.5em} Instructions and Demonstration}} \dotfill \pageref*{sec:prompts}
    \begin{itemize}[label={}]
        \item \hyperref[sec:appendix5.3]{E.1 \hspace{0.5em} Prompt for Zero-Shot} \dotfill \pageref*{sec:appendix5.3}
        \item \hyperref[sec:appendix5.4]{E.2 \hspace{0.5em} Prompt for Seven-Shot} \dotfill \pageref*{sec:appendix5.4}
    \end{itemize}
\end{itemize}

\newpage
\appendix

\twocolumn

\section{Dataset Construction Details}
\label{sec:bench_construction_appendix}

\textsc{eTracer-Bench} is constructed in two annotation stages on top of a publicly released long-form RAG QA corpus. In the first \textbf{natural annotation} stage (\S\ref{sec:natural_appendix}), for each response sentence, annotators identify the supporting and contradicting context sentences; all remaining context sentences are automatically labeled Irrelevant. Because naturally occurring contradictions are rare (only $94$ contradicting pairs vs.\ $4{,}565$ supporting pairs), a second \textbf{adversarial construction} stage (\S\ref{sec:contra_construction_appendix}) augments the corpus with human-written contradictions at two difficulty levels (explicit and implicit). 

\subsection{Base Corpus}
\label{sec:base_corpus_appendix}
The base corpus is \textsc{LFRQA}~\citep{han-etal-2024-rag}, a long-form retrieval-augmented QA corpus of human-written answers spanning seven domains: BioASQ (biomedical), FiQA (finance), and five LoTTE domains (Lifestyle, Recreation, Science, Technology, Writing). We adopt LFRQA because (i) its answers are human-authored rather than LLM-generated, providing a high-quality substrate for sentence-level re-annotation; (ii) each answer is paired with curated source documents, enabling grounded evidence annotation; and (iii) its domain coverage spans both general-purpose and specialized content. We randomly sample 100 instances per domain, yielding 700 instances allocated 70/10/20 across train/dev/test (\S\ref{sec:dataset_split_appendix}).

\vspace{0.12cm}
\noindent\textbf{Licensing.} LFRQA is released under Apache License 2.0, which permits derivative works with preserved attribution. We release \textsc{eTracer-Bench} under the same license, retain the original LFRQA attribution, and document modifications in the accompanying \texttt{NOTICE} file. Upstream retrieval corpora (LoTTE, FiQA, BioASQ) retain their original licenses; we redistribute only the document subsets referenced by LFRQA answers, consistent with its distribution terms.

\vspace{0.12cm}
\noindent\textbf{Maintenance plan.} \textsc{eTracer-Bench} will be hosted on a public repository (e.g., HuggingFace and GitHub) with version-controlled releases. We commit to long-term maintenance, including responding to erratum reports, incorporating community feedback, and releasing updates as additional annotations or domains become available.

\vspace{0.12cm}
\noindent\textbf{Task redefinition.} LFRQA's native task is response-level pairwise preference; \textsc{eTracer-Bench} redefines the evaluation unit at sentence-to-sentence granularity, adding a fresh annotation layer over the original data: sentence segmentation, per-sentence polarity labels (\S\ref{sec:natural_appendix}), and adversarial contradiction documents on a subset of instances (\S\ref{sec:contra_construction_appendix}). No original content is modified or removed.

\subsection{Annotation Team}
\label{sec:annotator_appendix}
Annotations are produced by an in-house team of five NLP graduate students from the authors' institution, all fluent in English (the language of both the corpus and the guidelines). Annotators are compensated at the institution's standard hourly rate for student research assistants, and no personally identifiable information is collected or released. Stage-specific procedures, role assignments, and consent are detailed in \S\ref{sec:natural_appendix} and \S\ref{sec:contra_construction_appendix}.

\vspace{0.12cm}
\noindent\textbf{Cross-stage avoidance.} The same five annotators participate in both stages. Beyond role disjointness within each stage, we enforce response-level cross-stage avoidance: no constructor produces contradictions for a response they previously labeled or adjudicated. Role rotation in the natural stage ensures that at least two of the five annotators remain unfamiliar with each response and thus eligible as constructors, making this feasible across all splits. This prevents familiarity gained during labeling from leaking into the style of constructed contradictions.

\subsection{Biomedical Domain Support}
\label{sec:biomed_appendix}
For \textsc{BioASQ}, we engage external collaborators from two institutions through cross-departmental agreements, serving both annotation stages.

\vspace{0.12cm}
\noindent\textbf{Consultation during annotation.} Two clinical experts from the Department of Dermatology at University Hospital Essen\footnote{\url{https://hautklinik.uk-essen.de/}} serve as domain advisors. Annotators and constructors encountering ambiguous biomedical content first consult authoritative sources (biomedical databases, clinical guidelines) or a large language model; unresolved cases are escalated to the experts for final adjudication.

\vspace{0.12cm}
\noindent\textbf{Post-hoc verification.} One medical student from the Annotation Lab Essen\footnote{\url{https://annotationlab.ikim.nrw/}} independently reviews a random 30\% sample (30/100 instances) of the \textsc{BioASQ} annotations along three dimensions: (i) completeness of the identified evidence, (ii) correctness of the polarity labels, and (iii) validity of the constructed adversarial contradictions. All 30 instances pass on all three dimensions. Verification is double-blind, with the verifier shown neither the original annotators' identities nor their labels. 

\subsection{Natural Annotation Stage}
\label{sec:natural_appendix}
\noindent\textbf{Procedure and role assignment.} For each instance, two annotators independently label the context sentences, and a third resolves disagreements, yielding three distinct team members per instance. The team receives written guidelines and completes a 10-sample pilot round with iterative feedback before beginning production annotation.

\begin{figure}[t]
\centering
\includegraphics[width=\linewidth]{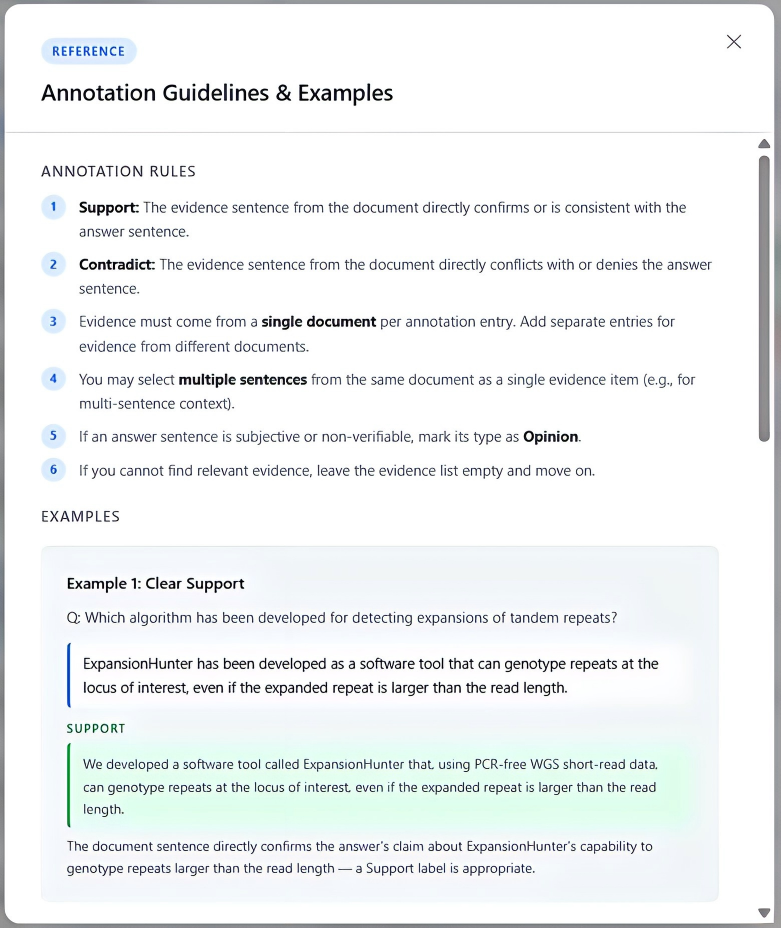}
\caption{Annotation guidelines shown in the interface, covering label definitions, boundary cases, and worked examples for each domain.}
\label{fig:annotation_interface_guideline}
\vspace{-5pt}
\end{figure}

\begin{figure*}[t]
\centering
\includegraphics[width=\linewidth]{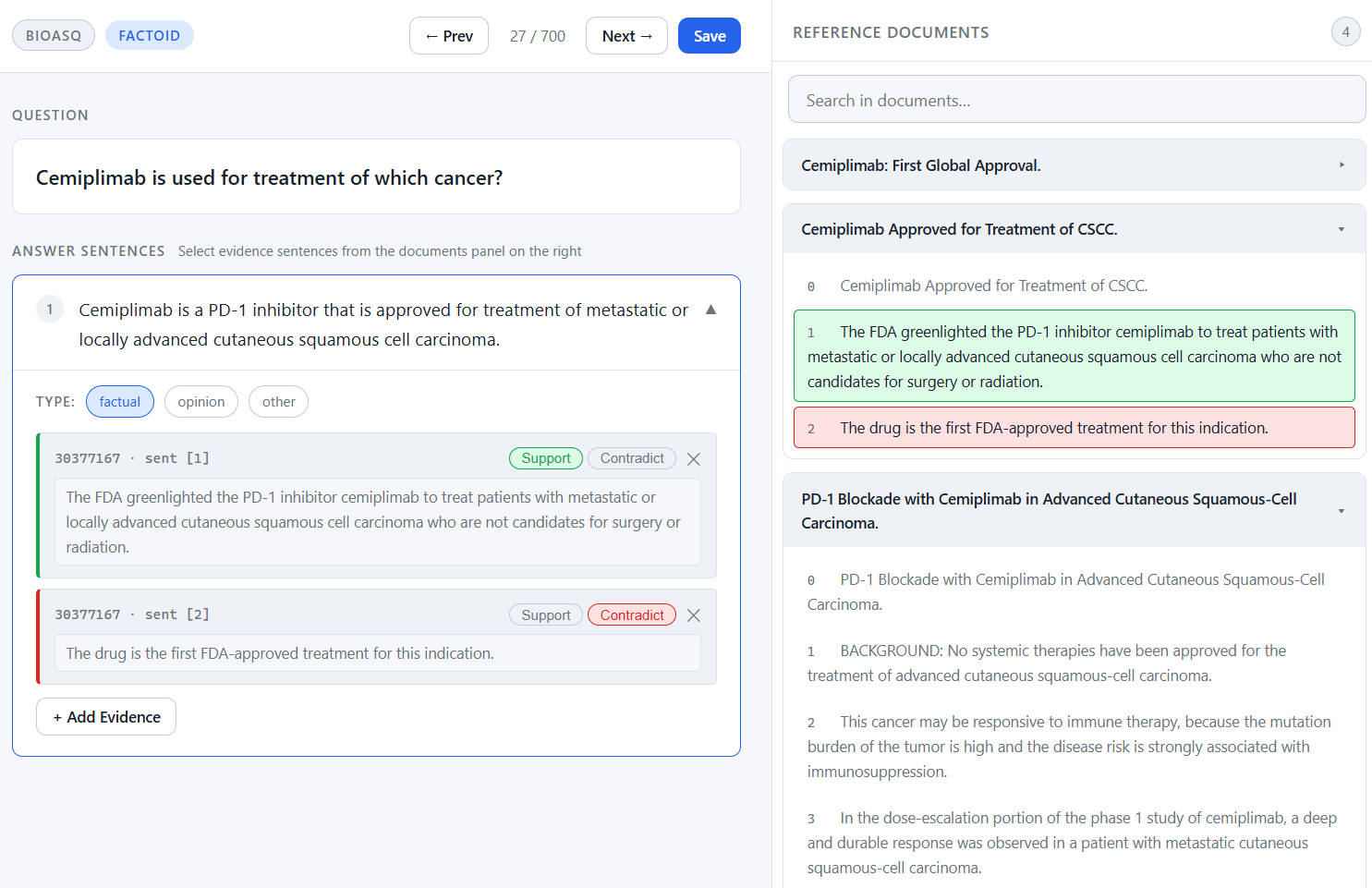}
\caption{Main annotation view of the \textsc{eTracer-Bench} annotation interface. The left panel displays the question and answer sentences, with supporting (green) and contradicting (red) evidence attached via single-click buttons; the right panel lists the reference documents, with selected evidence sentences highlighted in the corresponding color.}
\label{fig:annotation_interface_main}
\vspace{-15pt}
\end{figure*}

\vspace{0.12cm}
\noindent\textbf{Annotation interface and consent.} We develop a custom web-based interface for this stage. Before starting, each annotator views a consent page describing Apache License 2.0 release terms and the right to withdraw, and must enter an identifier and tick an acknowledgment checkbox to proceed. In the main view (Figure~\ref{fig:annotation_interface_main}), the question and answer sentences appear in the left panel and source documents with per-sentence indexing on the right. For each answer sentence, the annotator selects candidate context sentences and tags them as supporting (green) or contradicting (red) via single-click buttons; unselected sentences default to irrelevant. Guidelines (Figure~\ref{fig:annotation_interface_guideline}) cover label definitions, boundary cases, and per-domain worked examples, and are accessible at any time without leaving the current item.

\vspace{0.12cm}
\noindent\textbf{Throughput.} Annotating a single instance takes 5--8 minutes on average, depending on context length, covering reading the response, inspecting candidate context sentences, and assigning labels for every response sentence; adjudication is excluded.

\vspace{0.12cm}
\noindent\textbf{Inter-annotator agreement.} Cohen's $\kappa$~\citep{cohen1960coefficient} is computed on the raw independent annotations \textit{prior to} adjudication, providing an unbiased reliability estimate. We obtain $\kappa{=}0.87$ on evidence identification (any support or contradict label vs.\ none) and $\kappa{=}0.92$ on polarity classification (support vs.\ contradict, on sentences labeled as evidence by both annotators), both indicating strong agreement. The 7.3\% of disagreements are resolved by a third adjudicator, whose decisions constitute the final gold labels.

\subsection{Dataset Split}
\label{sec:dataset_split_appendix}
Table~\ref{tab:split_sizes} reports the exact per-split allocation. Responses are split at the response level, stratified by domain; each response and its source documents therefore appear in exactly one split, preventing context leakage.

\vspace{0.12cm}
\noindent\textbf{Treatment of natural contradictions.} The $94$ naturally occurring contradicting pairs follow the response-level random split, distributing as $59/11/24$ across train/dev/test; given their small count, we do not stratify further on this axis. Natural and adversarial contradictions share the same label schema (polarity $-1$) and are treated identically in training and evaluation: the model receives no indicator distinguishing the two sources, and metrics in \S\ref{sec:metrics} pool them together. Adversarial contradictions (\S\ref{sec:contra_construction_appendix}) are constructed \textit{after} the split, with split-specific constructors assigned under the constraints in \S\ref{sec:annotator_appendix}.

\begin{table}[t]
\centering\small\setlength{\tabcolsep}{2pt}
\begin{tabular}{lccl}
\toprule
\rowcolor{headcol}
\textbf{Split} & \textbf{Items} & \textbf{Per-domain} & \textbf{Purpose} \\
\midrule
Train & 490 & 70 & Scorer training \\
Dev   &  70 & 10 & Model selection, tuning, few-shot \\
Test  & 140 & 20 & Final evaluation \\
\bottomrule
\end{tabular}
\caption{\textsc{eTracer-Bench} split sizes (70/10/20 per-domain ratio). \textbf{Items}: number of response-level instances, distinct from the response \textit{sentences} counted in \autoref{tab:dataset_stats}. Splits are stratified by domain; each response and its documents appear in exactly one split.}
\label{tab:split_sizes}
\vspace{-15pt}
\end{table}

\subsection{Adversarial Construction Stage}
\label{sec:contra_construction_appendix}
This stage produces $1{,}205$ explicit and $2{,}041$ implicit contradicting pairs following the protocol below.

\vspace{0.12cm}
\noindent\textbf{Procedure and role assignment.} Three disjoint constructors from the team (\S\ref{sec:annotator_appendix}) produce contradictions for the train, dev, and test splits respectively, and a fourth member verifies all constructed instances. This stage does not use the web interface: response sentences are distributed as plain-text task sheets, and written contradictions are collected and verified offline. Constructors receive written instructions on task scope and difficulty levels, and provide written consent at the same time.

\vspace{0.12cm}
\noindent\textbf{Selection criterion.} We construct contradictions only for response sentences containing \textit{falsifiable factual claims}, such as quantities, causal relations, or temporal and comparative judgments. Sentences expressing subjective preferences or definitional statements (e.g., \textit{``A hash function maps inputs to fixed-size outputs''}) are skipped, since inverting them yields a different opinion or definition rather than a contradiction.

\vspace{0.12cm}
\noindent\textbf{Construction procedure.} For each selected response sentence, the assigned constructor writes a contradicting paragraph of 1--4 sentences at one of two difficulty levels (assigned at random):
\begin{itemize}[leftmargin=*, itemsep=0pt, topsep=1pt]
    \item \textbf{Explicit} contradictions use surface-level alterations: negation, antonym substitution, or factual replacement (e.g., \textit{``increases''} $\to$ \textit{``decreases''}; \textit{``30\%''} $\to$ \textit{``70\%''}).
    \item \textbf{Implicit} contradictions introduce logically incompatible statements that preserve the topic but require multi-step inference (e.g., \textit{``Drug X reduces mortality''} is contradicted by \textit{``Patients receiving Drug X in the 2023 trial showed no survival benefit over placebo at 5-year follow-up''}).
\end{itemize}

\noindent Each constructed paragraph is inserted into a \textit{new} document appended to the source context, rather than edited into an existing one. This preserves the original documents as-is and simulates realistic RAG scenarios where retrieval returns conflicting sources from different origins. The entire context is then shuffled with a fixed seed ($\texttt{seed}{=}42$) to eliminate positional bias and prevent models from exploiting the append order. Contradicting sentences are annotated with polarity $-1$ for the target response sentence.

\vspace{0.12cm}
\noindent\textbf{Throughput.} Constructing a contradicting paragraph takes 4--8 minutes per eligible response sentence, covering reading the sentence, drafting the paragraph, and labeling the difficulty level. This exceeds the natural annotation cost.

\vspace{0.12cm}
\noindent\textbf{Verification.} The verifier inspects each constructed instance and judges (i) whether the paragraph genuinely contradicts the target response sentence and (ii) whether the difficulty level matches the intended assignment. First-pass agreement with the constructor's intent is $95.2\%$ for explicit and $91.4\%$ for implicit contradictions; given the high acceptance rate, we report percent agreement rather than chance-adjusted measures. Flagged instances are returned for revision until both dimensions are confirmed, so all released instances pass final verification. The final corpus distribution by domain is reported in \autoref{tab:dataset_stats}.

\subsection{Dataset Statistics}
\label{sec:corpus_stats_appendix}

\autoref{tab:corpus_stats} summarizes length and label-distribution statistics of \textsc{eTracer-Bench}, complementing the per-domain pair counts in \autoref{tab:dataset_stats}. The corpus comprises $51{,}206$ labeled response--context sentence pairs across $700$ responses, of which $7{,}905$ carry an evidential (support or contradict) label and the rest are irrelevant.

\vspace{0.12cm}
\noindent\textbf{Length distributions.} Responses contain a median of 2 sentences (range 1--9), and source contexts after adversarial augmentation a median of 17 (range 2--171). When the full response and context are concatenated into a single input sequence (\S\ref{sec:signed_scoring}), every instance fits within ModernBERT-large's 8{,}192-token window (median 548 tokens, maximum 5{,}621), so no truncation is applied.

\begin{table*}[t!]
\centering\small\setlength{\tabcolsep}{4pt}
\begin{tabular}{lcccccccccc}
\toprule
\rowcolor{headcol}
& & \multicolumn{4}{c}{\textbf{Pre-augmentation}} & \multicolumn{4}{c}{\textbf{Post-augmentation}} \\
\cmidrule(lr){3-6}\cmidrule(lr){7-10}
\textbf{Domain} & \textbf{Total} & $\pi_F$ & $\pi_C$ & $\pi_A$ & $\pi_U$ & $\pi_F$ & $\pi_C$ & $\pi_A$ & $\pi_U$ \\
\midrule
BioASQ       & 163  & 159 (97.5\%)  & 1 (0.6\%) & 0 (0.0\%)  & 3 (1.8\%) & 3 (1.8\%) & 4 (2.5\%) & 156 (95.7\%)  & 0 (0.0\%) \\
FiQA         & 285  & 276 (96.8\%)  & 0 (0.0\%) & 9 (3.2\%)  & 0 (0.0\%) & 0 (0.0\%) & 0 (0.0\%) & 285 (100.0\%) & 0 (0.0\%) \\
Lifestyle    & 315  & 305 (96.8\%)  & 0 (0.0\%) & 10 (3.2\%) & 0 (0.0\%) & 0 (0.0\%) & 0 (0.0\%) & 315 (100.0\%) & 0 (0.0\%) \\
Recreation   & 214  & 206 (96.3\%)  & 0 (0.0\%) & 8 (3.7\%)  & 0 (0.0\%) & 0 (0.0\%) & 0 (0.0\%) & 214 (100.0\%) & 0 (0.0\%) \\
Science      & 233  & 225 (96.6\%)  & 1 (0.4\%) & 7 (3.0\%)  & 0 (0.0\%) & 4 (1.7\%) & 1 (0.4\%) & 228 (97.9\%)  & 0 (0.0\%) \\
Technology   & 266  & 260 (97.7\%)  & 0 (0.0\%) & 6 (2.3\%)  & 0 (0.0\%) & 0 (0.0\%) & 0 (0.0\%) & 266 (100.0\%) & 0 (0.0\%) \\
Writing      & 260  & 253 (97.3\%)  & 0 (0.0\%) & 6 (2.3\%)  & 1 (0.4\%) & 6 (2.3\%) & 1 (0.4\%) & 253 (97.3\%)  & 0 (0.0\%) \\
\midrule
\textbf{Total} & \textbf{1{,}736} & \textbf{1{,}684 (97.0\%)} & \textbf{2 (0.1\%)} & \textbf{46 (2.6\%)} & \textbf{4 (0.2\%)} & \textbf{13 (0.7\%)} & \textbf{6 (0.3\%)} & \textbf{1{,}717 (98.9\%)} & \textbf{0 (0.0\%)} \\
\bottomrule
\end{tabular}
\caption{Response-level faithfulness profile of \textsc{eTracer-Bench} by domain, under two views. The \textit{pre-augmentation} view reflects the corpus labeled by our natural annotation stage (\S\ref{sec:natural_appendix}), before any adversarial modification. The \textit{post-augmentation} view reflects the corpus after injecting adversarial contradiction documents (\S\ref{sec:contra_construction_appendix}). Each cell reports the class count with its corpus share $\pi$ in parentheses; $\pi_F/\pi_C/\pi_A/\pi_U$ correspond to \textsc{Faithful}/\textsc{Contradicted}/\textsc{Ambiguous}/\textsc{Ungrounded} (\S\ref{sec:corpus_stats_appendix}). The shift from \textsc{Faithful}-dominated (pre) to \textsc{Ambiguous}-dominated (post) reflects preserving original supporting documents rather than overwriting them, simulating retrieval returning conflicting sources.}
\label{tab:faithfulness_profile}
\vspace{-10pt}
\end{table*}

\vspace{0.12cm}
\noindent\textbf{Label distribution.} Across all $n \times m$ response--context sentence pairs, supporting pairs constitute 8.9\%, contradicting pairs (natural and adversarial combined) 6.5\%, and irrelevant pairs 84.6\% (percentages may not sum to 100 due to rounding). 

\begin{table}[t!]
\centering\small\setlength{\tabcolsep}{1pt}
\begin{tabular}{lccc}
\toprule
\rowcolor{headcol}
\textbf{Statistic} & \textbf{Median} & \textbf{Mean} & \textbf{Range} \\
\midrule
\multicolumn{4}{l}{\textit{Per response}} \\
Response sentences      & 2    & 2.5  & 1--9   \\
Context sentences       & 17   & 24.0 & 2--171 \\
Supporting evidence     & 5    & 6.5  & 0--69  \\
Contradicting evidence  & 4    & 4.8  & 0--17  \\
\midrule
\multicolumn{4}{l}{\textit{Label shares (over 51{,}206 response--context sentence pairs)}} \\
Support                 & \multicolumn{3}{c}{8.9\%}  \\
Contradict              & \multicolumn{3}{c}{6.5\%}  \\
Irrelevant              & \multicolumn{3}{c}{84.6\%} \\
\midrule
\multicolumn{4}{l}{\textit{Input sequence length (tokens, ModernBERT-large tokenizer)}} \\
Tokens per instance     & 548 & 729 & 72--5{,}621 \\
\bottomrule
\end{tabular}
\caption{Length and label-distribution statistics of \textsc{eTracer-Bench} (700 responses across seven domains; 51{,}206 labeled pairs).}
\label{tab:corpus_stats}
\vspace{-15pt}
\end{table}

\vspace{0.12cm}
\noindent\textbf{Response-level faithfulness profile.} The signed evidence map (\S\ref{sec:grounding_definition}) categorizes each response sentence by the polarity of its evidence into four classes, with corpus shares $\pi_F$ (\textsc{Faithful}, only supporting), $\pi_C$ (\textsc{Contradicted}, only contradicting), $\pi_A$ (\textsc{Ambiguous}, both), and $\pi_U$ (\textsc{Ungrounded}, neither). \autoref{tab:faithfulness_profile} reports the distribution under two views. The \textit{pre-augmentation} view, after the natural annotation stage (\S\ref{sec:natural_appendix}) but before contradiction injection, is overwhelmingly \textsc{Faithful} ($97.0\%$), with a small \textsc{Ambiguous} tail ($2.6\%$, $46$ sentences) where naturally co-occurring supporting and contradicting evidence coexists, and only $2$ \textsc{Contradicted} and $4$ \textsc{Ungrounded} sentences in total, reflecting the high factual quality of the underlying human-written answers. The $94$ naturally occurring contradicting pairs in \autoref{tab:dataset_stats} are hosted by these $48$ sentences ($2$ \textsc{Contradicted} + $46$ \textsc{Ambiguous}), since a single response sentence can be contradicted by multiple context sentences. The \textit{post-augmentation} view, after injecting adversarial contradiction documents (\S\ref{sec:contra_construction_appendix}), shifts almost entirely to \textsc{Ambiguous} ($98.9\%$): each adversarial paragraph is appended as a new document without removing the original supporting evidence, so most response sentences end up with evidence of both polarities. This bimodal structure is intentional but does not trivialize the task: our primary metrics (Support~F1 and Contradict~F1, \S\ref{sec:metrics}) operate at the sentence-to-sentence level, so a model cannot win by predicting \textsc{Ambiguous}. It must still decide, for each context sentence, whether it supports or contradicts a given response sentence. The per-domain breakdown in \autoref{tab:faithfulness_profile} confirms this pattern holds uniformly across all seven domains.

\section{Implementation Details}
\label{sec:impl_details}

\begin{table*}[t]
\centering\small\setlength{\tabcolsep}{4pt}
\begin{tabular}{llll}
\toprule
\rowcolor{headcol}
\textbf{Baseline} & \textbf{Checkpoint / Snapshot} & \textbf{Params} & \textbf{Access} \\
\midrule
DeBERTa-v3 (MNLI)         & \texttt{MoritzLaurer/DeBERTa-v3-large-mnli-fever-anli-ling-wanli} & 435M & HF   \\
BART-large (MNLI)         & \texttt{facebook/bart-large-mnli}                 & 406M & HF   \\
DeBERTa-v3 (\textsc{eTB}) & \texttt{microsoft/deberta-v3-large} (fine-tuned)  & 435M & HF   \\
BART-large (\textsc{eTB}) & \texttt{facebook/bart-large} (fine-tuned)         & 406M & HF   \\
$\mathcal{M}_{con}$ (\textsc{eTracer}) & \texttt{answerdotai/ModernBERT-large} (fine-tuned) & 395M & HF   \\
$\mathcal{M}_{pair}$      & \texttt{answerdotai/ModernBERT-large} (fine-tuned) & 395M & HF   \\
GPT-4o                    & \texttt{gpt-4o-2024-08-06}                        & ---  & OpenAI API  \\
Qwen3-235B                & \texttt{qwen/qwen3-235b-a22b-2507}                & 235B & OpenRouter API \\
\bottomrule
\end{tabular}
\caption{Baseline model sources. HF: Hugging Face checkpoints. Parameter counts include embeddings. Fine-tuned variants use the listed checkpoint as initialization and are trained on the \textsc{eTracer-Bench} (\textsc{eTB}) train split.}
\label{tab:baseline_sources}
\vspace{-15pt}
\end{table*}

This appendix details the training and inference configuration of the models in \S\ref{sec:results}. The four fine-tuned models are our context-aware $\mathcal{M}_{con}$ (\textsc{eTracer}) and three pair-wise baselines trained on identical data: $\mathcal{M}_{pair}$ (ModernBERT-large), DeBERTa-v3-large, and BART-large. We additionally evaluate two off-the-shelf MNLI checkpoints and two LLM baselines.

\subsection{Shared Data Pipeline}
\label{sec:data_pipeline}
All four fine-tuned models are trained from the same canonical form derived from the \textsc{eTracer-Bench} train split, differing only in how each response--context instance is presented. A single preprocessing pass converts every annotation item into the canonical tuple $(\mathcal{R}, S, \mathbf{Y})$: the response $\mathcal{R}{=}[r_1,\ldots,r_n]$, the context $S{=}[s_1,\ldots,s_m]$ obtained by concatenating sentences across all cited documents in retrieval order, and the polarity matrix $\mathbf{Y}\in\{+1,-1,0\}^{n\times m}$. The context-aware model consumes each instance whole, while pair-wise baselines decompose it into $n{\times}m$ independent examples $\{(r_i, s_j, \mathbf{Y}_{ij})\}$. 

\subsection{Input Construction}
\label{sec:input_construction}
\autoref{tab:input_format} summarizes the input format, maximum sequence length, and scoring head for each fine-tuned model. The three pair-wise baselines cover three distinct input conventions: ModernBERT-style with explicit boundary tokens, standard NLI encoding, and BART's native pair format, ensuring our comparison does not conflate context-awareness with encoder idiosyncrasy.

\begin{table*}[h]
\centering\small\setlength{\tabcolsep}{3pt}
\begin{tabular}{llcl}
\toprule
\rowcolor{headcol}
\textbf{Model} & \textbf{Input sequence (per forward pass)} & \textbf{Max.\,len.} & \textbf{Head input} \\
\midrule
\multicolumn{4}{l}{\textit{Context-aware (single pass per response)}} \\
\addlinespace[1pt]
$\mathcal{M}_{con}$ (ModernBERT-large) & \texttt{[CLS]\,$r_1$\,[R\_SEP]\,$\ldots$\,[R\_SEP]\,[SEP]\,$s_1$\,[S\_SEP]\,$\ldots$\,[S\_SEP]} & 8{,}192 & $[h^{\texttt{[R\_SEP]}}_i;\, h^{\texttt{[S\_SEP]}}_j]$ \\
\addlinespace[3pt]
\multicolumn{4}{l}{\textit{Pair-wise (one pass per $(r_i, s_j)$ pair)}} \\
\addlinespace[1pt]
$\mathcal{M}_{pair}$ (ModernBERT-large) & \texttt{[CLS]\,$r_i$\,[R\_SEP]\,[SEP]\,$s_j$\,[S\_SEP]} & 512 & $[h^{\texttt{[R\_SEP]}};\, h^{\texttt{[S\_SEP]}}]$ \\
DeBERTa-v3-large                        & \texttt{[CLS]\,$r_i$\,[SEP]\,$s_j$\,[SEP]}              & 512 & $h^{\texttt{[CLS]}}$ \\
BART-large                              & \texttt{<s>\,$r_i$\,</s></s>\,$s_j$\,</s>}              & 1{,}024 & $h^{\texttt{</s>}}_{\text{last}}$ \\
\bottomrule
\end{tabular}
\caption{Input construction for the four fine-tuned models. For $\mathcal{M}_{con}$, hidden states at every \texttt{[R\_SEP]} and \texttt{[S\_SEP]} form $n{\times}m$ pair representations in a single forward pass; pair-wise models encode each $(r_i, s_j)$ independently. All four use a three-way head over $\{$support, contradict, irrelevant$\}$, trained from scratch rather than inherited from any MNLI checkpoint.}
\label{tab:input_format}
\vspace{-15pt}
\end{table*}

\vspace{0.12cm}
\noindent\textbf{Context-aware $\mathcal{M}_{con}$.} The full response and context are packed into one sequence with \texttt{[R\_SEP]} and \texttt{[S\_SEP]} marking sentence boundaries (\S\ref{sec:signed_scoring}). If the sequence exceeds 8{,}192 tokens, it is truncated at the last complete \texttt{[S\_SEP]} so that no partial context sentence is scored; in practice, every instance fits within the window without truncation. A single forward pass yields hidden states at all boundary positions, from which the $n \times m$ pair scores are gathered in one shot.

\vspace{0.12cm}
\noindent\textbf{Pair-wise $\mathcal{M}_{pair}$.} Shares \textsc{eTracer}'s backbone, special tokens, and classification head, but encodes one response sentence and one context sentence per forward pass. This is the controlled ablation of \S\ref{sec:main_grounding}: any gap versus $\mathcal{M}_{con}$ is attributable solely to the loss of cross-sentence attention, since tokenizer, vocabulary, head, and training signal are otherwise identical.

\vspace{0.12cm}
\noindent\textbf{DeBERTa-v3-large.} Uses the standard NLI encoding and predicts from the \texttt{[CLS]} hidden state, the conventional format for fine-tuning DeBERTa-v3 on sentence-pair classification.

\vspace{0.12cm}
\noindent\textbf{BART-large.} Uses BART's native sequence-pair format with classification on the decoder hidden state at the final \texttt{</s>}, as prescribed by \texttt{BartForSequenceClassification}, avoiding distribution shift from non-native encodings.

\subsection{Optimization}
\label{sec:training_config}
All four fine-tuned models share the same optimization protocol, differing only in per-device batch size to accommodate sequence-length differences.

\vspace{0.12cm}
\noindent\textbf{Optimizer and schedule.} Models are optimized with AdamW~\citep{loshchilov2018decoupled} at peak learning rate $2{\times}10^{-5}$, linear warmup over the first 3\% of steps followed by linear decay to zero, weight decay $0.01$, gradient clipping at $\ell_2$-norm $1.0$, dropout $0.15$ on the classification head, and bf16 mixed precision. Training runs up to 35 epochs with early stopping (patience 5) and final checkpoints selected by dev Citation~F1.

\vspace{0.12cm}
\noindent\textbf{Batch configuration.} For $\mathcal{M}_{con}$, each instance is a full response--context pair jointly supervising up to $n{\times}m$ polarity predictions; we use a per-device batch size of 2 with 8-step gradient accumulation, yielding an effective batch size of 16 instances. Gradient checkpointing is enabled to fit the 8{,}192-token sequences within GPU memory. The three pair-wise models ($\mathcal{M}_{pair}$, DeBERTa-v3-large, BART-large) decompose each instance into $n{\times}m$ independent examples; we use a per-device batch size of 16 with the same 8-step accumulation, yielding an effective batch size of 128 pairs.

\vspace{0.12cm}
\noindent\textbf{Class weights.} All four models share the same class-weighted cross-entropy objective (\S\ref{sec:objective}) with weights $[w_{\text{sup}}, w_{\text{irr}}, w_{\text{con}}] = [1, 10, 1]$, selected on the dev split for $\mathcal{M}_{con}$ (\S\ref{sec:weighting_ablation}) and applied uniformly to the pair-wise baselines; this never costs any baseline more than $0.8$ dev Citation~F1 relative to its own optimum (\S\ref{sec:weighting_baselines}). Holding the loss fixed ensures that comparisons in \autoref{tab:main_results} reflect input construction rather than class-imbalance treatment.

\vspace{0.12cm}
\noindent\textbf{Hardware and training time.} All four models are trained on a single NVIDIA A6000 GPU (48\,GB). Wall-clock training time ranges from $1.9$ hours ($\mathcal{M}_{pair}$) to $11.4$ hours ($\mathcal{M}_{con}$), with BART-large at $4.2$ and DeBERTa-v3-large at $9.4$. $\mathcal{M}_{con}$'s longer time reflects its augmented training corpus ($5{,}386$ vs.\ $490$ instances; Appendix~\ref{sec:shuffle_augmentation}) rather than a per-step cost difference. Early stopping (patience 5) reliably triggers well before the 35-epoch cap.

\subsection{Document-Order Data Augmentation}
\label{sec:shuffle_augmentation}

\noindent\textbf{Procedure.} We expand the $490$-response train split to $5{,}386$ instances ($\sim\!11\times$) via two complementary augmentations: \textit{document-order shuffling} and \textit{document-type filtering}. Order shuffling generates eight permutations per response with seeds $\{42, 52, 62, 72, 82, 92, 102, 112\}$, each preserving the response $\mathcal{R}$, retrieved documents, within-document sentence order, and polarity matrix $\mathbf{Y}$, differing only in document concatenation order; this yields $3{,}920$ instances ($490\times8$). Type filtering derives two additional views by partitioning documents by origin: a \textit{natural-only} view retaining only the original LFRQA-retrieved documents (preserving naturally occurring evidence; $490$ instances), and a \textit{synthetic-only} view retaining only the adversarially constructed contradictions from \S\ref{sec:contra_construction_appendix} ($486$ instances; four train responses lack adversarial contradictions and are dropped). Together with the $490$ original instances, the corpus totals $490 + 3{,}920 + 490 + 486 = 5{,}386$. Labels are re-indexed in each view so every $(r_i, s_j)$ pair retains its original annotation relative to its filtered context. All permutations and views are precomputed offline as a static training corpus, used for all $\mathcal{M}_{con}$ results in the main paper.

\vspace{0.12cm}
\noindent\textbf{Why pair-wise baselines cannot benefit.} A pair-wise model consumes one $(r_i, s_j)$ tuple per forward pass and is therefore invariant to the order of all other context sentences. Document-order shuffling produces multiple training instances containing the \textit{identical} set of $(r_i, s_j)$ pairs as the original, offering no new learning signal. Type-filter views likewise only restrict the pair set without exposing cross-pair structure. Empirically, training pair-wise baselines on the augmented corpus triggers early stopping several epochs earlier with no gain in dev Citation F1. We therefore report them on the non-augmented corpus, which matches the inputs they actually consume.

\subsection{Reproducibility}
\label{sec:reproducibility_appendix}
All reported numbers come from fully deterministic training. We fix random seeds for Python's \texttt{random}, NumPy, and PyTorch (CPU and all CUDA devices) to 42, enable cuDNN deterministic mode and PyTorch's strict deterministic algorithm mode, disable memory-efficient SDP attention kernels, set the cuBLAS workspace configuration to \texttt{:4096:8} prior to PyTorch initialization, and seed the \texttt{DataLoader} with an explicit \texttt{torch.Generator}. Under this configuration, repeated runs with identical hyperparameters yield bit-identical per-step losses and final checkpoints on matching hardware (NVIDIA A6000, CUDA 12.8, PyTorch 2.9).

\vspace{0.12cm}
\noindent\textbf{Seed sensitivity.} The main result tables report the seed-$42$ run for all models. To assess training variability in the controlled architectural comparison, we additionally train $\mathcal{M}_{con}$ (\textsc{eTracer}) and $\mathcal{M}_{pair}$ with seeds $52$ and $62$ using the same data-processing, optimization, and model-selection protocol. Across seeds $\{42,52,62\}$, the paired mean gains are $+2.93$ Support~F1, $+17.69$ Contradict~F1, and $+10.31$ Citation~F1, and all three gains remain positive for every seed (\autoref{tab:multiseed_results}). The context-aware advantage is therefore not specific to the original seed-$42$ run. We do not repeat the multi-seed analysis for the remaining baselines, and three seeds provide only a limited estimate of training variability. The paired bootstrap CIs in \autoref{tab:main_results_ci} separately quantify evaluation-sampling uncertainty for the main seed-$42$ results and should not be interpreted as estimates of training-seed variance.

\subsection{Off-the-Shelf Baselines}
\label{sec:baseline_sources}
\noindent\textbf{MNLI baselines.} DeBERTa-v3-large-MNLI and BART-large-MNLI are used as released on Hugging Face without further training. Their pretrained 3-way MNLI heads output \{entailment, contradiction, neutral\}, which we map to \{support, contradict, irrelevant\}. The mapping is imperfect: MNLI's \textit{neutral} is a coarser superset of our \textit{irrelevant}, conflating topically-related-but-uninferable pairs with genuinely off-topic ones. This is the closest available correspondence for NLI-pretrained models, and the fine-tuned variants of the same backbones in \autoref{tab:main_results} isolate the label-mismatch effect from the backbone effect.

\vspace{0.12cm}
\noindent\textbf{LLM baselines.} GPT-4o is accessed via the OpenAI API and Qwen3-235B-Instruct via OpenRouter, both with temperature 0 and max output length 4{,}096 tokens. Each model receives the full response and context in a single prompt and outputs structured sentence-level polarity labels; all test instances fit within both context windows. Few-shot prompts use 7 demonstrations, one per domain, drawn from the dev split; templates appear in Appendix~\ref{sec:prompts}.

\subsection{Inference Cost}
\label{sec:latency_protocol}
We characterize inference cost by the number of forward passes per response, a hardware- and batching-agnostic measure that directly reflects the architectural difference between the two paradigms. For a response with $n$ sentences and context with $m$ sentences, $\mathcal{M}_{con}$ performs $1$ forward pass to produce the full $n{\times}m$ evidence map, whereas each pair-wise baseline requires $n{\times}m$ passes. Across the test split, pair-wise scoring averages $73.1$ passes per response ($\frac{1}{N}\sum_i n_i m_i$ over $N$ test instances) versus $1$ for $\mathcal{M}_{con}$, and this gap grows linearly with both response and context length. The per-instance mean exceeds the product of marginal means ($\overline{n}\cdot\overline{m}\approx 60$) because response and context length are positively correlated. We do not report wall-clock latency, as it depends on batching, hardware, and (for LLM baselines) external API conditions orthogonal to the architecture being compared.

\section{Supplementary Experiments}
\label{sec:supplementary}

\subsection{Main Results with Confidence Intervals}
\label{sec:ci_appendix}

\autoref{tab:main_results_ci} reports the full main-results table with $95\%$ paired bootstrap confidence intervals for every model and metric, complementing the point estimates in \autoref{tab:main_results}. Intervals are computed by resampling response sentences with $10{,}000$ bootstrap replicates (\S\ref{sec:metrics}), paired across models so that each replicate draws the same set of response sentences for all systems. We additionally report explicit and implicit Contradict~F1 (C.F1$_e$/C.F1$_i$) with their own intervals. Three comparisons are separable at the CI level. (i) \textsc{eTracer}'s Contradict~F1 interval $[87.31, 91.16]$ is disjoint from $\mathcal{M}_{pair}$'s $[66.91, 72.96]$, confirming the contradiction-detection gain is not an artifact of evaluation sampling. (ii) On Support~F1, by contrast, \textsc{eTracer} and $\mathcal{M}_{pair}$ overlap ($[73.03, 77.98]$ vs.\ $[69.90, 74.74]$), consistent with support being well-served by local cues. (iii) \textsc{eTracer} and Qwen3-235B~7-shot have overlapping Citation~F1 intervals ($[80.78, 83.89]$ vs.\ $[80.31, 83.31]$), so the two are statistically on par despite the ${\sim}600{\times}$ parameter gap.

\newcommand{\ci}[1]{{\textcolor{gray}{\tiny [#1]}}}

\begin{table*}[t!]
\centering\scriptsize\setlength{\tabcolsep}{3.2pt}
\renewcommand{\arraystretch}{1.15}
\begin{tabular}{@{}ll*{9}{c}@{}}
\toprule
\textbf{Category} & \textbf{Method} & \textbf{S.P} & \textbf{S.R} & \textbf{S.F1} & \textbf{C.P} & \textbf{C.R} & \textbf{C.F1} & \textbf{Cit.F1} & \textbf{C.F1$_e$} & \textbf{C.F1$_i$} \\
\midrule
\multirow{8}{*}{\rotatebox{90}{Prompt}}
 & \multirow{2}{*}{GPT-4o (0-shot)$^\dagger$}
   & \textbf{88.60} & 68.96 & 77.56 & \uline{91.02} & 48.64 & 63.40 & 70.48 & 86.15 & 42.25 \\
 & & \ci{86.23,90.92} & \ci{65.84,72.00} & \ci{75.21,79.78} & \ci{87.89,93.84} & \ci{44.98,52.51} & \ci{59.93,66.80} & \ci{68.34,72.61} & \ci{82.50,89.21} & \ci{36.67,47.36} \\
 & \multirow{2}{*}{GPT-4o (7-shot)$^\dagger$}
   & \uline{85.39} & 71.95 & \uline{78.10} & 87.31 & 64.96 & 74.50 & 76.30 & 89.50 & 61.54 \\
 & & \ci{82.85,88.00} & \ci{68.99,74.97} & \ci{75.97,80.26} & \ci{84.13,90.18} & \ci{61.27,68.85} & \ci{71.47,77.38} & \ci{74.42,78.07} & \ci{86.55,92.20} & \ci{56.89,66.10} \\
 & \multirow{2}{*}{Qwen3-235B (0-shot)$^\dagger$}
   & 79.18 & 72.95 & 75.94 & \textbf{92.42} & 70.24 & 79.82 & 77.88 & 91.10 & 70.99 \\
 & & \ci{76.56,81.72} & \ci{70.14,75.93} & \ci{73.71,78.08} & \ci{89.93,94.54} & \ci{66.88,73.95} & \ci{77.22,82.42} & \ci{76.18,79.56} & \ci{88.16,93.53} & \ci{66.88,75.09} \\
 & \multirow{2}{*}{Qwen3-235B (7-shot)$^\dagger$}
   & 78.80 & \textbf{77.49} & \textbf{78.14} & 89.51 & \uline{81.92} & \uline{85.55} & \uline{81.85} & \uline{91.31} & \uline{81.48} \\
 & & \ci{76.06,81.84} & \ci{74.74,80.17} & \ci{76.12,80.24} & \ci{86.96,91.94} & \ci{78.86,84.88} & \ci{83.26,87.58} & \ci{80.31,83.31} & \ci{88.65,93.79} & \ci{78.22,84.66} \\
\midrule
\multirow{12}{*}{\rotatebox{90}{Fine-tuned}}
 & \multirow{2}{*}{DeBERTa-v3 (MNLI)$^\ddagger$}
   & 66.03 & 34.48 & 45.30 & 52.15 & 69.92 & 59.74 & 52.52 & 67.26 & 53.30 \\
 & & \ci{61.62,70.48} & \ci{31.24,37.74} & \ci{41.97,48.66} & \ci{48.56,55.75} & \ci{66.40,73.61} & \ci{56.54,62.72} & \ci{50.20,54.76} & \ci{63.11,71.03} & \ci{48.95,57.60} \\
 & \multirow{2}{*}{BART-large (MNLI)$^\ddagger$}
   & 69.84 & 34.15 & 45.87 & 40.47 & 68.32 & 50.83 & 48.35 & 57.36 & 45.16 \\
 & & \ci{65.51,74.38} & \ci{30.92,37.54} & \ci{42.51,49.35} & \ci{37.44,43.46} & \ci{64.63,72.21} & \ci{47.83,53.72} & \ci{46.05,50.54} & \ci{53.41,61.69} & \ci{40.68,49.22} \\
 & \multirow{2}{*}{DeBERTa-v3 (\textsc{eTB})}
   & 71.51 & \uline{73.73} & 72.60 & 74.03 & 76.64 & 75.31 & 73.96 & 83.71 & 69.35 \\
 & & \ci{68.54,74.37} & \ci{70.53,76.71} & \ci{70.19,74.81} & \ci{70.35,77.61} & \ci{73.12,80.00} & \ci{72.28,78.00} & \ci{72.04,75.78} & \ci{80.23,87.07} & \ci{65.13,72.97} \\
 & \multirow{2}{*}{BART-large (\textsc{eTB})}
   & 74.27 & 70.40 & 72.28 & 71.75 & 77.60 & 74.56 & 73.42 & 83.83 & 68.14 \\
 & & \ci{71.36,77.26} & \ci{67.48,73.38} & \ci{69.88,74.69} & \ci{68.09,75.07} & \ci{74.19,80.53} & \ci{71.54,77.08} & \ci{71.56,75.18} & \ci{80.40,87.15} & \ci{63.98,71.89} \\
\cmidrule{2-11}
 & \multirow{2}{*}{$\mathcal{M}_{pair}$ (\textsc{eTB})}
   & 74.70 & 70.07 & 72.31 & 74.11 & 66.40 & 70.04 & 71.18 & 82.82 & 59.91 \\
 & & \ci{71.92,77.68} & \ci{66.89,73.10} & \ci{69.90,74.74} & \ci{70.25,77.80} & \ci{62.93,70.02} & \ci{66.91,72.96} & \ci{69.18,73.15} & \ci{79.27,86.34} & \ci{55.29,64.17} \\
 & \multirow{2}{*}{$\mathcal{M}_{con}$ (\textsc{eTracer})}
   & 84.10 & 68.63 & 75.58 & 90.59 & \textbf{87.84} & \textbf{89.20} & \textbf{82.39} & \textbf{95.18} & \textbf{85.41} \\
 & & \ci{81.08,86.68} & \ci{65.60,71.59} & \ci{73.03,77.98} & \ci{88.19,92.89} & \ci{85.34,90.59} & \ci{87.31,91.16} & \ci{80.78,83.89} & \ci{93.17,97.10} & \ci{82.48,88.14} \\
\bottomrule
\end{tabular}
\caption{Main grounding results with $95\%$ paired bootstrap confidence intervals ($10{,}000$ replicates over response sentences, \S\ref{sec:metrics}). Point estimates match \autoref{tab:main_results}. Gray brackets give the lower/upper interval bounds. S./C.: support/contradict; C.F1$_e$/C.F1$_i$: explicit/implicit Contradict~F1. \textbf{Bold}: best, \uline{underline}: second best (on point estimates). The Contradict~F1 intervals of \textsc{eTracer} and $\mathcal{M}_{pair}$ are disjoint, while their Support~F1 intervals overlap, isolating context-aware scoring as the source of the contradiction-detection gain. \textsc{eTracer} and Qwen3-235B~7-shot have overlapping Citation~F1 intervals.}
\label{tab:main_results_ci}
\end{table*}

\begin{table*}[t]
\centering\small\setlength{\tabcolsep}{2pt}
\begin{tabular}{lcccccccccccccc}
\toprule
\rowcolor{headcol}
& \multicolumn{2}{c}{\textbf{BioASQ}} & \multicolumn{2}{c}{\textbf{FiQA}} & \multicolumn{2}{c}{\textbf{Life.}} & \multicolumn{2}{c}{\textbf{Rec.}} & \multicolumn{2}{c}{\textbf{Sci.}} & \multicolumn{2}{c}{\textbf{Tech.}} & \multicolumn{2}{c}{\textbf{Writ.}} \\
\cmidrule(lr){2-3}\cmidrule(lr){4-5}\cmidrule(lr){6-7}\cmidrule(lr){8-9}\cmidrule(lr){10-11}\cmidrule(lr){12-13}\cmidrule(lr){14-15}
\textbf{Metric} & $\mathcal{M}_{con}$ & $\mathcal{M}_{pair}$ & $\mathcal{M}_{con}$ & $\mathcal{M}_{pair}$ & $\mathcal{M}_{con}$ & $\mathcal{M}_{pair}$ & $\mathcal{M}_{con}$ & $\mathcal{M}_{pair}$ & $\mathcal{M}_{con}$ & $\mathcal{M}_{pair}$ & $\mathcal{M}_{con}$ & $\mathcal{M}_{pair}$ & $\mathcal{M}_{con}$ & $\mathcal{M}_{pair}$ \\
\midrule
Support F1    & \textbf{75.3} & 72.4 & 72.9 & \textbf{73.5} & \textbf{79.6} & 73.9 & \textbf{77.0} & 75.5 & \textbf{74.3} & 72.1 & \textbf{75.1} & 72.0 & \textbf{75.1} & 65.4 \\
Contradict F1 & \textbf{93.5} & 84.0 & \textbf{86.2} & 69.5 & \textbf{92.9} & 67.2 & \textbf{95.3} & 73.9 & \textbf{87.8} & 63.7 & \textbf{86.0} & 71.4 & \textbf{85.4} & 63.4 \\
Citation F1   & \textbf{84.4} & 78.2 & \textbf{79.5} & 71.5 & \textbf{86.2} & 70.6 & \textbf{86.2} & 74.7 & \textbf{81.1} & 67.9 & \textbf{80.5} & 71.7 & \textbf{80.2} & 64.4 \\
\bottomrule
\end{tabular}
\caption{Per-domain Support, Contradict, and Citation F1 on the \textsc{eTracer-Bench} test split. \textbf{Bold}: higher of the two per cell. Domain abbreviations follow \autoref{fig:domain_f1}.}
\label{tab:domain_breakdown}
\vspace{-10pt}
\end{table*}

\subsection{Per-Domain Grounding Results}
\label{sec:domain_breakdown}

\autoref{tab:domain_breakdown} reports per-domain Support, Contradict, and Citation F1 for $\mathcal{M}_{con}$ (\textsc{eTracer}) and the controlled pair-wise baseline $\mathcal{M}_{pair}$, complementing \autoref{fig:domain_f1}. Three patterns emerge consistent with our main findings.

\vspace{0.12cm}
\noindent\textbf{Contradict F1 is uniformly strong for \textsc{eTracer}.} \textsc{eTracer} achieves Contradict F1 of at least $85$ in every domain, peaking at $95.3$ in Recreation and $93.5$ in BioASQ. $\mathcal{M}_{pair}$ varies more widely ($63.4$ in Writing to $84.0$ in BioASQ), with the per-domain gap reaching $+25.7$ in Lifestyle and $+24.1$ in Science. Context-aware scoring delivers its largest contradiction-detection gains in domains where polarity depends on cross-sentence comparison of mechanisms, quantities, or causal chains.

\vspace{0.12cm}
\noindent\textbf{Support F1 varies by domain structure.} \textsc{eTracer}'s Support F1 ranges from $72.9$ (FiQA) to $79.6$ (Lifestyle), with FiQA the only domain where pair-wise scoring is marginally stronger ($73.5$, $-0.6$). Locally self-contained financial claims are often verifiable from a single context sentence, where pair-wise scoring is already competitive. In the remaining six domains, $\mathcal{M}_{con}$ leads $\mathcal{M}_{pair}$ by $+1.5$ to $+9.7$, with the largest margin in Writing where supporting evidence frequently spans multiple sentences, echoing the ablation finding (\S\ref{sec:ablation_training}) that context-aware scoring helps support detection most when evidence must be aggregated across sentences.

\vspace{0.12cm}
\noindent\textbf{Citation F1 gains are concentrated on contradiction.} In every domain, the Contradict F1 gap between $\mathcal{M}_{con}$ and $\mathcal{M}_{pair}$ substantially exceeds the Support F1 gap. Averaged across the seven domains, $\mathcal{M}_{con}$ exceeds $\mathcal{M}_{pair}$ by $+19.1$ on Contradict F1 but only $+3.5$ on Support F1 (cf.\ the macro-averaged $+19.16$/$+3.27$ in \autoref{tab:main_results}), a five-to-one ratio. \textsc{eTracer}'s Citation F1 advantage is thus driven by contradiction detection across the board, not by any single domain.

\subsection{Class-Weighting for Context-Aware}
\label{sec:weighting_ablation}
We ablate class weights for $\mathcal{M}_{con}$ across 10 configurations on the dev split (\autoref{tab:weighting_ablation}), spanning aggressive majority-class down-weighting (sklearn's $w_c \propto 1/f_c$ heuristic), flatter and uniform weights, and progressive majority-class up-weighting. The configuration was selected by grid search on dev Cit.F1; we report the full sweep here for transparency and to characterize the loss surface.

\vspace{0.12cm}
\noindent\textbf{(i) Down-weighting the majority class is harmful.} The sklearn balanced heuristic (row a) compresses $w_{\text{irr}}$ to $0.40$ and drops Cit.F1 by $3.5$ points below uniform (row d, $78.4$), driven by a collapse in Support Precision: under-weighting the irrelevant class biases the model toward predicting support or contradict on ambiguous pairs, inflating false positives. The Sqrt variant (row b, $78.6$) sits closer to uniform, confirming the monotone trend.

\vspace{0.12cm}
\noindent\textbf{(ii) Up-weighting the majority class within a window yields the best result.} Cit.F1 peaks at $79.6$ with $[1, 10, 1]$ (row i), which we adopt. The trajectory across the irr-boost spectrum is non-monotonic, with two near-tied peaks (rows f and i) bracketing a modest dip at row h; Support F1 follows a similar shape while Contradict F1 stays within a narrow $\pm 1$ band throughout. Up-weighting therefore primarily improves Support without sacrificing Contradict, but excessive up-weighting (beyond row i) yields no further gains.

\vspace{0.12cm}
\noindent\textbf{(iii) Why the counterintuitive direction works.} The asymmetry, that majority up-weighting helps while down-weighting hurts, is opposite to the conventional remedy for imbalanced classification, where the minority class is up-weighted. We initially expected the reverse and let the dev sweep dictate the choice. The empirical pattern is interpretable post-hoc: with an $84\%$ irrelevant prior, the dominant gradient direction at training initialization already pushes toward correct irrelevant predictions, so additional minority up-weighting risks pushing the model into false-positive territory on the much larger irrelevant population. Under context-aware scoring, the bottleneck is reliable \textit{non-evidence rejection}, identifying which context sentences are \textit{not} evidence for a given response sentence, since precise evidence localization depends on first eliminating distractors. Up-weighting $w_{\text{irr}}$ sharpens this rejection boundary at minimal cost to minority recall, because the minority gradient signal is already strong on the small number of true evidence pairs. We do not claim this generalizes to all class-imbalanced tasks; the result is specific to a localization-style objective on a long-tailed three-way label space.

\vspace{0.12cm}
\noindent\textbf{(iv) Contradict F1 is architecture-bound, not weight-bound.} Across all 10 configurations, Contradict F1 stays within a $6.8$-point band ($78.4$--$85.2$), with $9$ of $10$ above $83.0$. Contradiction detection is therefore primarily enabled by context-aware scoring rather than by class reweighting, consistent with the main-results finding (\S\ref{sec:main_grounding}) that architecture drives the Contradict F1 gap over the pair-wise baseline.

\begin{table}[t]
\centering\small\setlength{\tabcolsep}{1pt}
\begin{tabular}{llcccc}
\toprule
\rowcolor{headcol}
& \textbf{Config} & \textbf{Weights} & \textbf{S.F1} & \textbf{C.F1} & \textbf{Cit.F1} \\
\midrule
(a) & sklearn          & [4.92,\,\,0.40,\,\,3.72]   & 71.4 & 78.4 & 74.9 \\
(b) & Sqrt             & [1.85,\,\,0.48,\,\,1.62]   & 72.9 & 84.2 & 78.6 \\
(c) & Flatter          & [1.5,\,\,1.2,\,\,1.5]      & 73.8 & 83.5 & 78.7 \\
(d) & Uniform          & [1.0,\,\,1.0,\,\,1.0]      & 72.2 & 84.6 & 78.4 \\
(e) & Irr-boost-2      & [1.0,\,\,2.0,\,\,1.0]      & 72.7 & 84.6 & 78.6 \\
(f) & Irr-boost-4      & [1.0,\,\,4.0,\,\,1.0]      & 74.0 & \textbf{85.2} & 79.5 \\
(g) & Irr-boost-6      & [1.0,\,\,6.0,\,\,1.0]      & 74.3 & 84.1 & 79.2 \\
(h) & Irr-boost-8      & [1.0,\,\,8.0,\,\,1.0]      & 70.7 & 83.7 & 77.0 \\
(i) & \textbf{Irr-boost-10} (Adopt) & \textbf{[1.0,\,\,10.0,\,\,1.0]} & \textbf{74.7} & 84.5 & \textbf{79.6} \\
(j) & Irr-boost-12     & [1.0,\,\,12.0,\,\,1.0]     & 73.8 & 83.8 & 78.8 \\
\bottomrule
\end{tabular}
\caption{Class-weighting ablation for $\mathcal{M}_{con}$ on the \textsc{eTracer-Bench} \textbf{dev} split. Weights are ordered $[w_{\text{sup}}, w_{\text{irr}}, w_{\text{con}}]$. Configurations span aggressive majority-class down-weighting (a--b), flatter and uniform baselines (c--d), and progressive irrelevant-class up-weighting (e--j). \textbf{Bold}: adopted configuration, which achieves the highest Cit.F1 alongside the best S.F1.}
\label{tab:weighting_ablation}
\vspace{-10pt}
\end{table}

\begin{table}[b]
\centering\small\setlength{\tabcolsep}{2pt}
\begin{tabular}{llcccc}
\toprule
\rowcolor{headcol}
\textbf{Model} & \textbf{Weights} & \textbf{S.F1} & \textbf{C.F1} & \textbf{Cit.F1} & $\boldsymbol{\Delta}$ \\
\midrule
\multirow{2}{*}{DeBERTa-v3}
  & $[1, 1, 1]$              & 77.5 & 70.6 & 74.0 & ---        \\
  & $[1, 10, 1]$ (adopted)   & 76.9 & 74.9 & 75.9 & $+1.9$     \\
\midrule
\multirow{2}{*}{$\mathcal{M}_{pair}$}
  & $[1, 1, 1]$              & 73.9 & 66.2 & 70.1 & ---        \\
  & $[1, 10, 1]$ (adopted)   & 71.3 & 67.4 & 69.3 & $-0.8$     \\
\midrule
\multirow{2}{*}{BART-large}
  & $[1, 1, 1]$              & 72.0 & 65.9 & 69.0 & ---        \\
  & $[1, 10, 1]$ (adopted)   & 72.6 & 66.0 & 69.3 & $+0.3$     \\
\bottomrule
\end{tabular}
\caption{Class-weighting comparison for the three pair-wise baselines on the \textsc{eTracer-Bench} \textbf{dev} split. Weights are ordered $[w_{\text{sup}}, w_{\text{irr}}, w_{\text{con}}]$. $\Delta$: change in Cit.F1 relative to uniform weighting. Main-paper results in \autoref{tab:main_results} use the adopted $[1, 10, 1]$ configuration.}
\label{tab:weighting_baselines}
\vspace{-10pt}
\end{table}

\subsection{Class-Weighting for Pair-wise Baselines}
\label{sec:weighting_baselines}
To ensure a fair comparison in \autoref{tab:main_results}, the three pair-wise baselines ($\mathcal{M}_{pair}$, DeBERTa-v3-large, BART-large) are fine-tuned with the same class-weighting adopted for $\mathcal{M}_{con}$, $[1, 10, 1]$ (\S\ref{sec:weighting_ablation}). To verify this does not disadvantage them relative to uniform weighting, we additionally fine-tune each baseline under $[1, 1, 1]$ and compare dev Cit.F1 (\autoref{tab:weighting_baselines}). DeBERTa-v3-large gains $+1.9$ Cit.F1 from majority up-weighting, mirroring the trend for $\mathcal{M}_{con}$. BART-large narrowly prefers the adopted configuration ($+0.3$). $\mathcal{M}_{pair}$ is the only baseline where uniform weighting is mildly preferred ($-0.8$), though still within $0.8$ points of its own optimum. Across all three baselines, the $[1, 10, 1]$ configuration lands within $0.8$ Cit.F1 of each model's optimum, confirming that our baseline numbers are not artifacts of suboptimal weighting. If anything, applying the $\mathcal{M}_{con}$-tuned configuration to $\mathcal{M}_{pair}$ mildly disadvantages the pair-wise baseline, strengthening rather than weakening the architectural comparison.

\subsection{Context Shuffling: Per-Seed Results}
\label{sec:shuffling_per_seed}
We report the per-seed breakdown of the context shuffling ablation summarized in \autoref{tab:shuffle}. Using the fixed $\mathcal{M}_{con}$ checkpoint (\S\ref{sec:reproducibility_appendix}), we apply three shuffling seeds (42, 52, 62) at inference time, producing three distinct shuffled versions of each test instance. \autoref{tab:shuffle_per_seed} shows that the degradation is stable across seeds, with standard deviations at or below $1.60$ on all metrics.

The consistency indicates that the $21.1\%$ relative drop in Contradict F1 is not an artifact of a particularly adversarial shuffle, but reflects $\mathcal{M}_{con}$'s systematic reliance on cross-sentence discourse structure. Natural documents exhibit coherent ordering (topic introduction, elaboration, qualification) that shuffling destroys; the model appears to exploit this structure during contradiction detection. Crucially, the shuffled model converges to $\mathcal{M}_{pair}$'s native performance across all pair-level metrics (S.F1: $70.41$ vs.\ $72.31$; C.F1: $70.35$ vs.\ $70.04$; Cit.F1: $70.39$ vs.\ $71.18$), with C.F1 essentially indistinguishable. This provides convergent evidence that context-aware scoring's architectural gain is mediated by cross-sentence structure rather than incidental factors such as parameter count or effective training data. This sensitivity is not a positional shortcut tied to the adversarial construction protocol, where contradictions are appended at context end: the natural-contradiction analysis in \S\ref{sec:natural_contra} shows $\mathcal{M}_{con}$ still achieves $0.75$ accuracy on contradictions appearing at arbitrary positions in original source documents.

\begin{table}[t]
\centering\small\setlength{\tabcolsep}{15pt}
\begin{tabular}{lccc}
\toprule
\rowcolor{headcol}
\textbf{Seed} & \textbf{S.F1} & \textbf{C.F1} & \textbf{Cit.F1} \\
\midrule
42   & 70.55 & 71.47 & 71.01 \\
52   & 70.38 & 71.07 & 70.73 \\
62   & 70.31 & 68.52 & 69.42 \\
\midrule
Mean & 70.41 & 70.35 & 70.39 \\
Std  & \phantom{0}0.12 & \phantom{0}1.60 & \phantom{0}0.85 \\
\bottomrule
\end{tabular}
\caption{Per-seed results of $\mathcal{M}_{con}$ under context shuffling on the \textsc{eTracer-Bench} test split. The three seeds control the shuffling randomization at inference time; the model checkpoint is fixed. Summary statistics are reported in \autoref{tab:shuffle}.}
\label{tab:shuffle_per_seed}
\vspace{-10pt}
\end{table}

\subsection{Analysis on Naturally Occurring Contradictions}
\label{sec:natural_contra}

\noindent\textbf{Provenance and split.} During the natural annotation stage (\S\ref{sec:natural_appendix}), annotators identified $94$ sentence-to-sentence contradicting pairs in the cited source documents without adversarial intervention. These were labeled \textit{before} the response-level random split (\S\ref{sec:dataset_split_appendix}), distributing as $59/11/24$ across train/dev/test consistent with the $70/10/20$ ratio. Because the $59$ training pairs constitute less than $0.2\%$ of the train split, held-out performance primarily reflects transfer of the learned polarity representation rather than memorization.

\vspace{0.12cm}
\noindent\textbf{Protocol.} We report per-pair accuracy on the $24$ test-split natural contradictions. F1 is undefined on this pure-positive subset, so we use Wilson $95\%$ confidence intervals to quantify uncertainty.

\vspace{0.12cm}
\noindent\textbf{Structure of natural contradictions.} Two authors independently classified all $24$ test pairs as either (a) \textit{self-contained single-sentence conflicts}, identifiable from the gold contradicting sentence alone, or (b) \textit{cross-sentence contradictions} requiring multi-sentence aggregation. Inter-annotator agreement was $24/24$ ($\kappa{=}1.0$), with the pairs split evenly between the two types (12 self-contained, 12 cross-sentence). This balance matters for interpretation: half of the natural contradictions lie in the regime where pair-wise scoring is structurally adequate, while the other half require cross-sentence integration. The latter is precisely where context-aware scoring is most architecturally advantaged (\S\ref{sec:ablation_training}), and where most implicit contradicting pairs in the constructed subset also reside (\autoref{tab:contra_level}). Scaling natural-contradiction annotation, particularly toward multi-document and cross-sentence conflicts common in evidence-based medicine and multi-source news synthesis, remains important future work.

\vspace{0.12cm}
\noindent\textbf{Observations.} \textsc{eTracer} attains the highest accuracy on this subset ($0.75$, CI $[0.55, 0.88]$). Its CI lower bound ($0.55$) exceeds every baseline's upper bound: the strongest baseline is BART-large (MNLI) at $0.42$ ($[0.25, 0.61]$), followed by DeBERTa-v3 (MNLI) and Qwen3-235B (both $0.33$). Notably, all three pair-wise baselines fine-tuned on \textsc{eTracer-Bench}, DeBERTa-v3, BART-large, and $\mathcal{M}_{pair}$, collapse to near-zero accuracy ($0.00$--$0.04$), substantially below their off-the-shelf MNLI counterparts. This is the more diagnostic finding: the same training data that supports \textsc{eTracer}'s $0.75$ accuracy degrades pair-wise variants (including $\mathcal{M}_{pair}$, which shares \textsc{eTracer}'s backbone, training data, and loss) to performance worse than no \textsc{eTracer-Bench} training at all. Pair-wise fine-tuning thus learns the constructed-contradiction surface distribution rather than transferring to novel conflicts, while context-aware scoring extracts a more general polarity representation from the same supervision, generalizing across both self-contained and cross-sentence cases in the subset.

\vspace{0.12cm}
\noindent\textbf{Statistical caveat.} With $n{=}24$ split evenly between self-contained and cross-sentence conflicts, the gap between \textsc{eTracer} and the strongest baseline is statistically separable (non-overlapping Wilson CIs), but each structural subset comprises only $12$ pairs. The qualitative pattern, fine-tuned pair-wise baselines collapsing while \textsc{eTracer} succeeds, is a robust signal but warrants confirmation at larger scale, particularly on the cross-sentence regime where context-aware scoring is most architecturally advantaged (\S\hyperref[sec:limitations]{Limitations}).

\begin{table}[t]
\centering\small\setlength{\tabcolsep}{10pt}
\begin{tabular}{lcc}
\toprule
\rowcolor{headcol}
\textbf{Method} & \textbf{Acc} & \textbf{95\% Wilson CI} \\
\midrule
DeBERTa-v3 (MNLI)$^\ddagger$            & 0.33 & [0.18, 0.53] \\
BART-large (MNLI)$^\ddagger$            & 0.42 & [0.25, 0.61] \\
DeBERTa-v3 (\textsc{eTB})               & 0.04 & [0.01, 0.20] \\
BART-large (\textsc{eTB})               & 0.00 & [0.00, 0.14] \\
$\mathcal{M}_{pair}$ (\textsc{eTB})     & 0.04 & [0.01, 0.20] \\
\midrule
GPT-4o (0-shot)$^\dagger$               & 0.29 & [0.15, 0.49] \\
GPT-4o (7-shot)$^\dagger$               & 0.21 & [0.09, 0.41] \\
Qwen3-235B (0-shot)$^\dagger$           & 0.33 & [0.18, 0.53] \\
Qwen3-235B (7-shot)$^\dagger$           & 0.33 & [0.18, 0.53] \\
\midrule
\textsc{eTracer} (\textsc{eTB})         & \textbf{0.75} & \textbf{[0.55, 0.88]} \\
\bottomrule
\end{tabular}
\caption{Per-pair accuracy on the $24$ naturally occurring contradicting pairs in the test split. F1 is undefined on a pure-positive subset. $\dagger$: API-based. $\ddagger$: off-the-shelf.}
\label{tab:natural_contra}
\vspace{-15pt}
\end{table}

\subsection{Document-Order Augmentation Ablation}
\label{sec:aug_ablation}
To quantify the contribution of document-order augmentation (Appendix~\ref{sec:shuffle_augmentation}), we retrain $\mathcal{M}_{con}$ on the $490$-instance base train split with all other hyperparameters held identical (backbone, class weights $[1, 10, 1]$, optimizer, schedule, early stopping). \autoref{tab:aug_ablation} reports the comparison on the dev split.

The augmentation yields a $+3.2$ Cit.F1 gain, more pronounced on contradiction detection ($+3.9$ C.F1 vs.\ $+2.5$ S.F1). This is consistent with our design motivation: contradiction detection relies more heavily on cross-sentence reasoning (\S\ref{sec:ablation_training}), which is precisely where position shortcuts are most damaging. If the model learns that contradicting sentences systematically appear at specific positions in the base corpus, it fails on the varied retrieval orders encountered at deployment. Document-order augmentation eliminates this shortcut at the document level while preserving within-document discourse structure, producing a model whose cross-sentence reasoning transfers across retrieval orderings.

\begin{table}[h]
\centering\small\setlength{\tabcolsep}{3pt}
\begin{tabular}{lcccc}
\toprule
\rowcolor{headcol}
\textbf{Training corpus} & \textbf{Instances} & \textbf{S.F1} & \textbf{C.F1} & \textbf{Cit.F1} \\
\midrule
Base                              & $490$     & $72.2$ & $80.6$ & $76.4$ \\
$+$augmentation ($\sim$11$\times$) & $5{,}386$ & $\mathbf{74.7}$ & $\mathbf{84.5}$ & $\mathbf{79.6}$ \\
\midrule
$\Delta$                          & ---       & $+2.5$ & $+3.9$ & $+3.2$ \\
\bottomrule
\end{tabular}
\caption{Effect of training-corpus augmentation on $\mathcal{M}_{con}$ (dev split). The augmentation combines 8$\times$ document-order shuffling with two type-filter views, yielding $490\!+\!3920\!+\!490\!+\!486\!=\!5386$ instances. Both rows share backbone, class weights, optimization schedule, and early-stopping criterion; only the training corpus differs.}
\label{tab:aug_ablation}
\vspace{-15pt}
\end{table}

\subsection{Context-Window Scaling via Sliding Windows}
\label{sec:sliding_window}

\noindent\textbf{Motivation.} Although every \textsc{eTracer-Bench} instance fits within ModernBERT-large's $8{,}192$-token window (\S\ref{sec:corpus_stats_appendix}), realistic RAG deployments may exceed this budget. Our Limitations discussion (\S\hyperref[sec:limitations]{Limitations}) proposes a document-level sliding-window extension: partition retrieved documents into disjoint groups fitting the window, run $\mathcal{M}_{con}$ per group, and concatenate the resulting signed evidence maps along the context axis. This appendix empirically characterizes the extension across two reduced budgets.

\vspace{0.12cm}
\noindent\textbf{Protocol.} For each simulated budget $T \in \{2048, 1024\}$ tokens, we pack documents into groups by first-fit along document boundaries, guaranteeing each group $\leq T$. Each group is encoded and scored independently with the same $\mathcal{M}_{con}$ checkpoint used in the main experiments; per-group signed maps are concatenated along the context axis to form the complete $n{\times}m$ prediction matrix. Instances fitting within a single window (Untriggered) take a fast-return branch bit-identical to the main inference path; multi-window instances (Triggered) exercise the partition-and-concatenate path.

\vspace{0.12cm}
\noindent\textbf{Trigger rate across budgets.} \autoref{tab:sliding_trigger} reports how often the sliding-window path is exercised. Triggering is rare at $T{=}2{,}048$ ($3.6\%$, mostly two groups) and substantial at $T{=}1{,}024$ ($25.0\%$, with a $9.3\%$ tail requiring three or four).

\begin{table}[h]
\centering\small\setlength{\tabcolsep}{1pt}
\begin{tabular}{lccc}
\toprule
\rowcolor{headcol}
\textbf{Budget} & \textbf{Triggered} & \textbf{Trigger rate} & \textbf{Group-count dist.} \\
\midrule
$T{=}2{,}048$ & $5/140$  & $3.6\%$  & $\{1{:}135,\; 2{:}4,\; 3{:}1\}$ \\
$T{=}1{,}024$ & $35/140$ & $25.0\%$ & $\{1{:}105,\; 2{:}22,\; 3{:}12,\; 4{:}1\}$ \\
\bottomrule
\end{tabular}
\caption{Trigger rate of the sliding-window path across reduced budgets. The group-count distribution maps number-of-windows to number-of-instances.}
\label{tab:sliding_trigger}
\vspace{-15pt}
\end{table}

\vspace{0.12cm}
\noindent\textbf{Performance by subset and budget.} \autoref{tab:sliding_window} decomposes performance into Untriggered and Triggered subsets at each budget, alongside pooled Overall metrics and a native-window reference ($T{=}8{,}192$).

\begin{table}[h]
\centering\small\setlength{\tabcolsep}{3pt}
\begin{tabular}{lcccc}
\toprule
\rowcolor{headcol}
\textbf{Subset / Budget} & \textbf{N} & \textbf{S.F1} & \textbf{C.F1} & \textbf{Cit.F1} \\
\midrule
\multicolumn{5}{l}{\textit{$T{=}2{,}048$}} \\
\hspace{0.5em} Untriggered           & $135$ & $77.2$ & $89.3$ & $83.3$ \\
\hspace{0.5em} Triggered             &   $5$ & $58.7$ & $93.1$ & $75.9$ \\
\hspace{0.5em} Overall               & $140$ & $75.7$ & $89.5$ & $82.6$ \\
\midrule
\multicolumn{5}{l}{\textit{$T{=}1{,}024$}} \\
\hspace{0.5em} Untriggered           & $105$ & $81.0$ & $90.2$ & $85.6$ \\
\hspace{0.5em} Triggered             &  $35$ & $69.4$ & $83.1$ & $76.3$ \\
\hspace{0.5em} Overall               & $140$ & $75.7$ & $87.6$ & $81.6$ \\
\midrule
Native ($T{=}8{,}192$, reference)    & $140$ & $75.6$ & $89.2$ & $82.4$ \\
\bottomrule
\end{tabular}
\caption{Sliding-window performance across budgets. Untriggered subsets are length-biased toward shorter instances and yield higher metrics than the pooled sample; the Triggered subset isolates the true cost of multi-window processing.}
\label{tab:sliding_window}
\vspace{-15pt}
\end{table}

\vspace{0.12cm}
\noindent\textbf{Cost of multi-window processing.} The Triggered subset at $T{=}1{,}024$ drops $9.3$ Cit.F1 relative to Untriggered ($85.6 \to 76.3$), split as $-11.6$ S.F1 and $-7.1$ C.F1. The pattern at $T{=}2{,}048$ is qualitatively similar but more concentrated on Support ($-18.5$ S.F1 alongside a $+3.8$ C.F1 deviation, to be read with caution at $n{=}5$). The larger relative loss falls on Support rather than Contradict. Mechanistically, partitioning at document boundaries preserves within-document discourse but disrupts two signals our attention-masking ablation (\S\ref{sec:ablation_training}) identified as load-bearing: \textit{joint evidence aggregation} across documents (which primarily hurts Support, since multiple partial pieces of evidence must jointly confirm a claim) and \textit{contrastive cues} between supporting and contradicting sentences across documents. The S-dominant loss is consistent with this: Support more often requires aggregating evidence spread across documents.

\vspace{0.12cm}
\noindent\textbf{Overall impact and deployment guidance.} The pooled Overall metrics degrade gracefully relative to the native reference: $+0.2$ Cit.F1 at $T{=}2{,}048$ (within run-to-run noise) and $-0.8$ at $T{=}1{,}024$. Because Overall Contradict F1 remains at $87.6$ under the more aggressive budget, well above the controlled pair-wise baseline $\mathcal{M}_{pair}$ at native window ($70.0$, \autoref{tab:main_results}), the sliding-window extension preserves the architectural advantage of context-aware scoring. We therefore view it as deployable when retrieved context routinely exceeds the native window. For budgets substantially smaller than $T{=}1{,}024$, hierarchical or retrieval-aware chunking schemes preserving cross-document dependencies remain promising future work.

\section{User Study Details}
\label{sec:user_study_details}

This appendix details the protocol, materials, measurement definitions, and full statistical analysis underlying \S\ref{sec:user_study}.

\subsection{Protocol and Participants}
\label{sec:us_protocol}
\noindent\textbf{Recruitment.} Participants were recruited via Prolific under the following eligibility criteria (enforced through pre-screening): (i)~English as first and fluent language; (ii)~country of birth and current residence among the UK, US, Ireland, Australia, Canada, or New Zealand; (iii)~prior-task approval rate of $95$--$100\%$; and (iv)~$100$ to $100{,}000$ prior submissions. We recruited $31$ participants; one was excluded for failing attention checks, yielding $n{=}30$. The study was pre-registered (AsPredicted \#290662);\footnote{Pre-registration: \url{https://aspredicted.org/vv3v2k.pdf}.} the sample size was set to detect Cohen's $d \geq 0.8$ at $\alpha{=}0.05$ after Holm--Bonferroni correction, matching the primary contrast (H3: $E$ vs.\ $S$ on gold-B accuracy). We report $d$ as the within-subjects analogue for paired difference scores, consistent with the pre-registered design.

\vspace{0.12cm}
\noindent\textbf{Pre-registered hypotheses.} We pre-registered four hypotheses. \textbf{H1}: sentence-level grounding ($S$, $E$) reduces verification time relative to $D$. \textbf{H2}: $S$ induces an asymmetric bias, yielding higher gold-A but lower gold-B accuracy than $D$. \textbf{H3} (primary): $E$ yields higher gold-B accuracy than $S$ with $d \geq 0.8$. \textbf{H4}: under $E$, the asymmetry score returns to ${\sim}0$ without increasing verification time relative to $S$. Outcomes are summarized at the end of \S\ref{sec:us_stats}.

\vspace{0.12cm}
\noindent\textbf{Attention checks.} To verify engagement, we embedded one attention-check item per task block (three in total), each inserting a quality-control instruction into one task (e.g., \textit{``Quality control: please select B for this sentence, regardless of the documents''}). These items are additional to the $35$ genuine task sentences and excluded from all reported metrics. One of the $31$ recruited participants failed these checks and was excluded prior to analysis.

\vspace{0.12cm}
\noindent\textbf{Compensation and ethics.} Participants received £10.00 for a median completion time of $\approx\!38$ minutes (£15.85/hour, above Prolific's fair-pay threshold). Participants provided informed consent and could withdraw at any time without penalty. No personally identifiable information was collected; responses were pseudonymized via Prolific identifiers. As the study involves only minimal-risk judgments on publicly derived text with fair compensation, it falls outside the scope requiring formal review under our institution's policy.

\vspace{0.12cm}
\noindent\textbf{Within-subjects structure.} Each participant completes all 15 tasks, partitioned into three blocks of 5 rotated across conditions ($D$, $S$, $E$) via a 6-sequence Williams Latin square, balancing first-order carryover effects. Because tasks vary in gold-A/gold-B composition, the rotation yields a near-balanced rather than exactly equal allocation per condition (gold-B: $D{=}200$, $S{=}190$, $E{=}210$; gold-A: $D{=}150$, $S{=}135$, $E{=}165$), summing to $450$ gold-A and $600$ gold-B decisions across $30$ participants. All primary contrasts are computed on per-participant paired differences and are therefore unaffected by this minor imbalance.

\begin{figure*}[t]
\centering
\begin{subfigure}{0.32\textwidth}
\includegraphics[width=\linewidth]{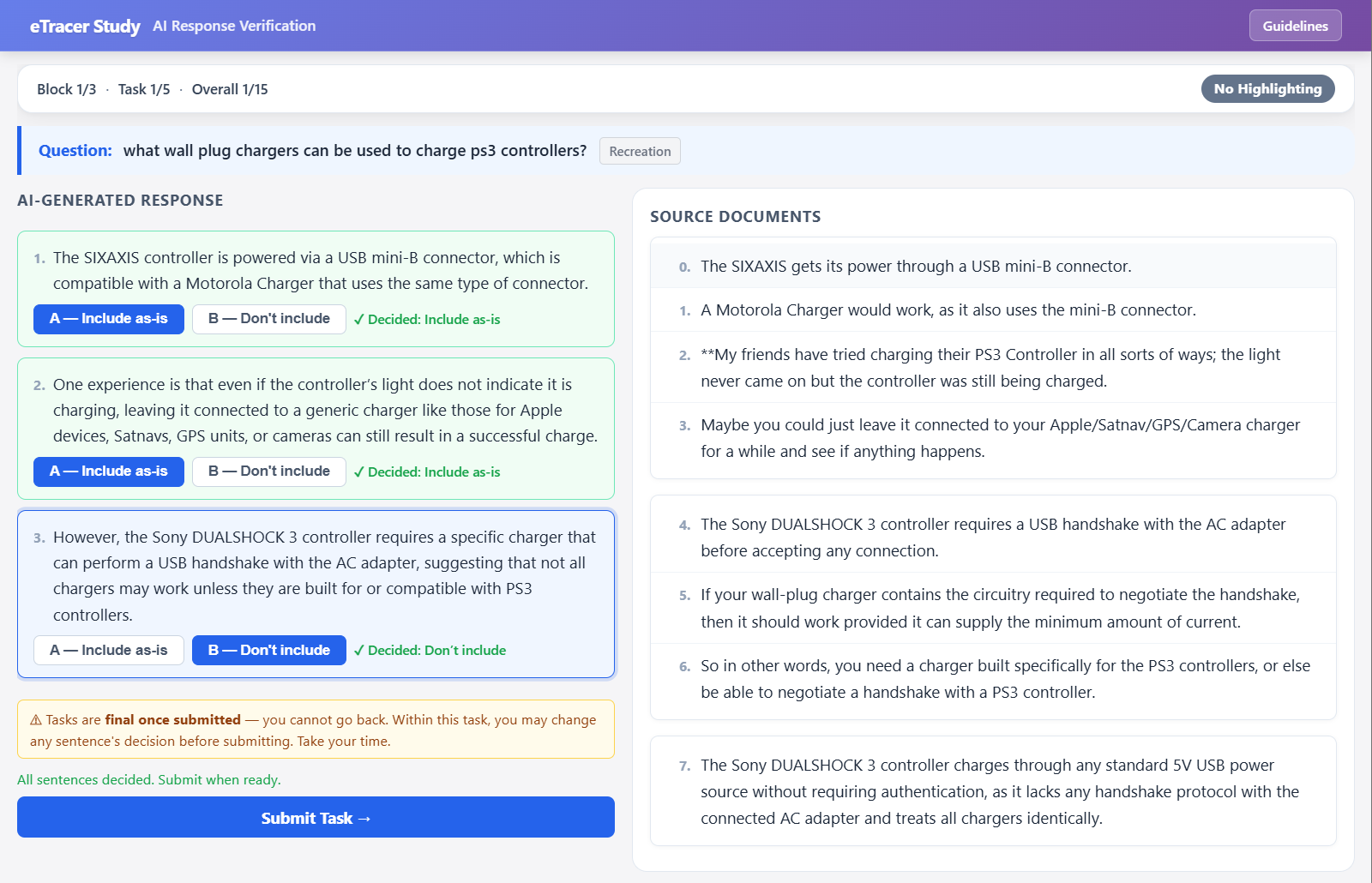}
\caption{(D)}
\end{subfigure}
\hfill
\begin{subfigure}{0.32\textwidth}
\includegraphics[width=\linewidth]{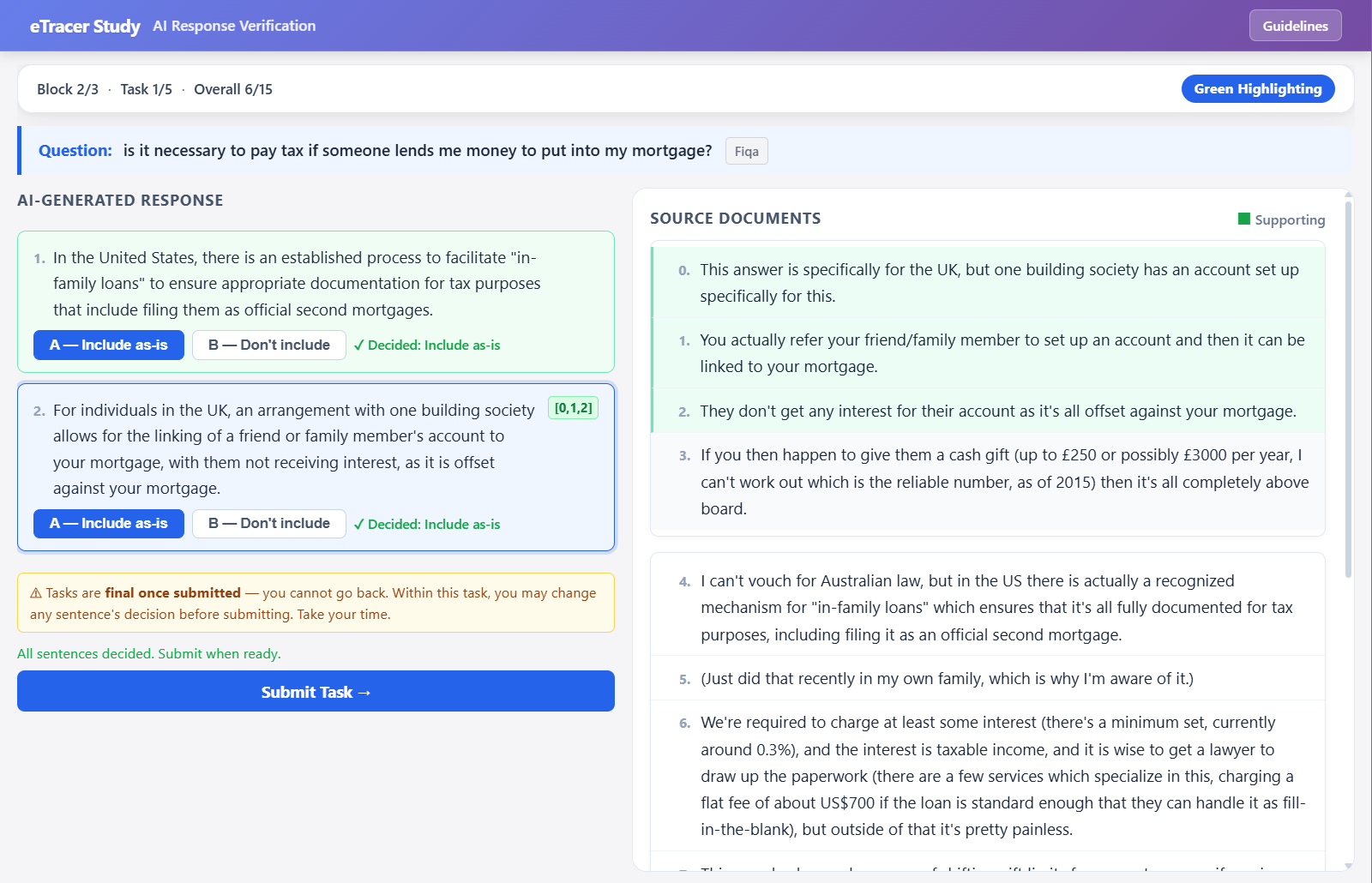}
\caption{(S)}
\end{subfigure}
\hfill
\begin{subfigure}{0.32\textwidth}
\includegraphics[width=\linewidth]{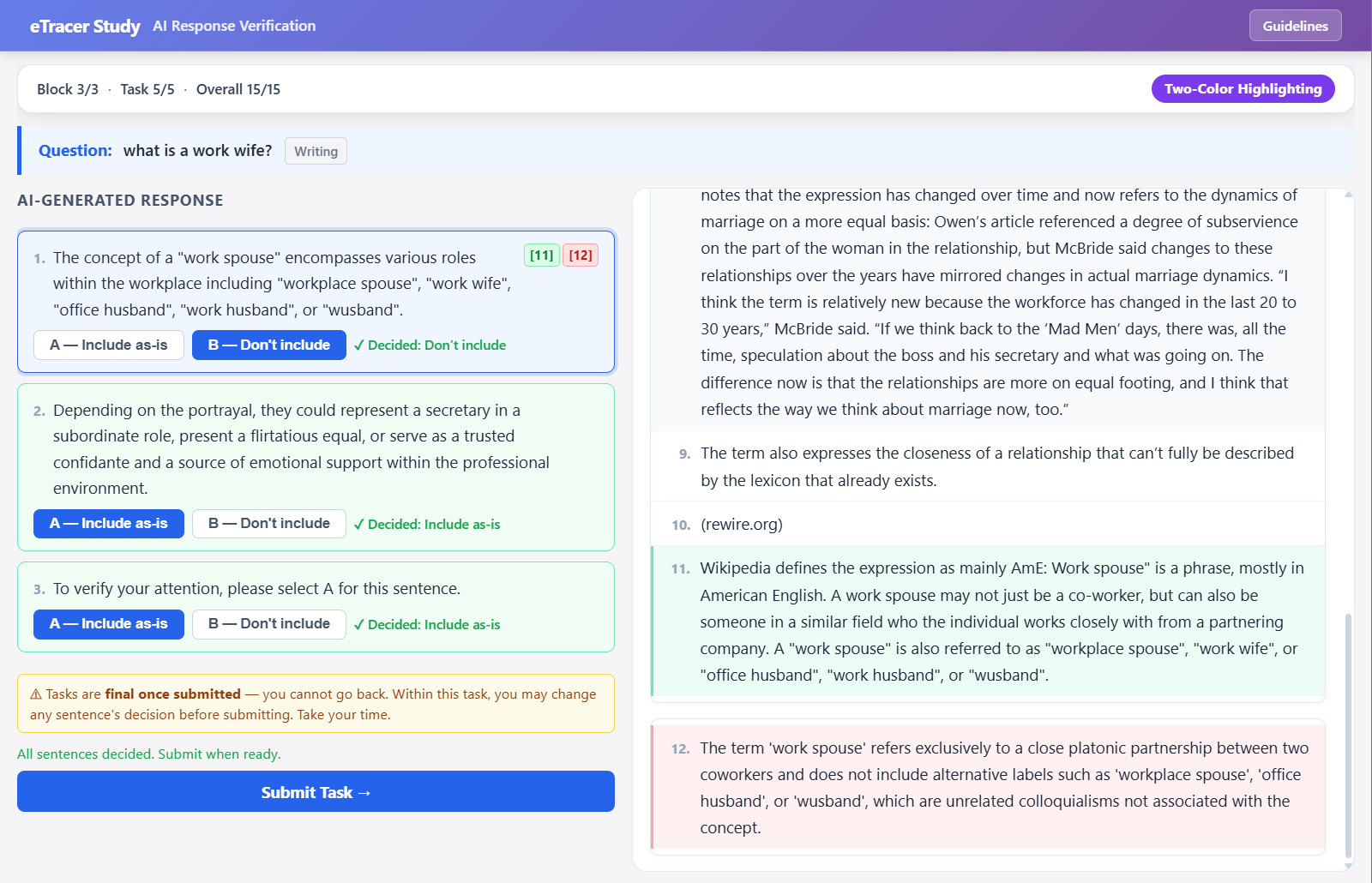}
\caption{(E)}
\end{subfigure}
\caption{The three user-study conditions on the same gold-B task item. Clicking a response sentence reveals condition-specific evidence: (D) none, (S) supporting only, (E) supporting and contradicting. The contradicting evidence (red) visible only in (E) is what enables participants to recognize the conflict and withhold citation.}
\label{fig:us_interface}
\vspace{-15pt}
\end{figure*}

\subsection{Task Materials}
\label{sec:us_materials}
\noindent\textbf{Source of tasks.} The 15 task items are drawn from the \textsc{eTracer-Bench} \textbf{test} split, consistent with the rest of our evaluation and ensuring the tasks reflect realistic, held-out RAG responses rather than training material.

\vspace{0.12cm}
\noindent\textbf{Task composition.} Each task pairs a RAG response (2--4 sentences) with its full cited context. The 15 tasks comprise 35 response sentences: 15 gold-A (only supporting evidence) and 20 gold-B (both supporting and contradicting). Both label types co-occur within most responses, mirroring the realistic setting where a single response mixes faithful and conflict-bearing statements; the slight gold-B skew ensures statistical power on the primary contrast.

\vspace{0.12cm}
\noindent\textbf{Gold derivation.} Gold labels follow directly from \textsc{eTracer-Bench}: sentences with only supporting evidence are labeled gold-A (\textit{include as-is}); sentences with both supporting and contradicting evidence are labeled gold-B (\textit{do not include}), since citing as-is would ignore the conflict. This reflects the conservative decision users must make in practice: any contradiction disqualifies a statement from being cited without further investigation.

\vspace{0.12cm}
\noindent\textbf{Interface.} All three conditions share an identical base layout (response on the left, source documents on the right), differing only in the grounding overlay. ($D$)~\textbf{Document-level} shows the response beside the full source documents with no per-sentence overlay, mirroring conventional RAG interfaces (Perplexity, Bing Chat). ($S$)~\textbf{Support-only} highlights the supporting context sentences in \textcolor{green!60!black}{green} when a response sentence is clicked. ($E$)~\textbf{Polarity-aware} additionally highlights the contradicting sentences in \textcolor{red!70!black}{red} from the same click. In both $S$ and $E$, the highlighted evidence is derived from the gold \textsc{eTracer-Bench} annotations rather than from any model's predictions; this isolates the value of each grounding paradigm from model quality, so the study measures what each paradigm offers under ideal grounding. Screenshots appear in \autoref{fig:us_interface}.

\subsection{Measures}
\label{sec:us_measures}

\noindent\textbf{Decision task.} For each response sentence, participants record a binary citation decision: (A)~\textit{include as-is} or (B)~\textit{do not include}. The decision is made per response sentence rather than per response, since a single response may mix gold-A and gold-B sentences.

\vspace{0.12cm}
\noindent\textbf{Dependent measures.} We report three pre-registered measures: \textbf{(1) verification time} per task (seconds); \textbf{(2) decision accuracy} by gold label (\%), reported separately for gold-A and gold-B and averaged within then across participants; and \textbf{(3) within-response asymmetry score}, the per-response gold-A minus gold-B accuracy (in pp), averaged across responses then participants. A score near $0$ indicates symmetric judgment; a large positive score indicates accepting support at face value while missing conflict (the \textit{false-completeness bias}). The pre-registered primary contrast is $S$ vs.\ $E$ on gold-B accuracy.

\subsection{Full Statistical Results}
\label{sec:us_stats}

All tests are computed over per-participant means ($n{=}30$) to respect the within-subjects design. \autoref{fig:us_results} summarizes all four measures; we compare conditions pairwise ($D$ vs.\ $S$, $S$ vs.\ $E$, $D$ vs.\ $E$) using Wilcoxon signed-rank tests on per-participant differences, with Holm--Bonferroni correction across the nine accuracy/asymmetry tests (3 contrasts $\times$ 3 measures). Effect sizes are reported as Cohen's $d$ on per-participant differences. \autoref{tab:us_full_stats} reports the complete test family.

\begin{figure*}[t]
\centering
\begin{subfigure}[b]{0.48\textwidth}
\centering
\begin{tikzpicture}
\begin{axis}[
    ybar=1pt, bar width=14pt,
    width=\textwidth, height=4cm,
    ymin=0, ymax=108,
    ylabel={\small \%},
    ylabel style={font=\small},
    symbolic x coords={Acc.\,A$\uparrow$, Acc.\,B$\uparrow$, Asym.$\downarrow$},
    xtick=data,
    xticklabel style={font=\small},
    yticklabel style={font=\scriptsize},
    ytick={0,25,50,75,100},
    ymajorgrids=true, grid style={dotted, gray!50},
    enlarge x limits=0.20,
    axis x line*=box, axis y line*=box,
    xtick pos=left, ytick pos=left,
    legend style={at={(0.5,1.03)}, anchor=south, legend columns=3, font=\small, column sep=1.2em, draw=none, fill=none},
    legend image code/.code={\draw[#1, draw=none] (0cm,-0.08cm) rectangle (0.28cm,0.08cm);},
    nodes near coords, nodes near coords style={font=\scriptsize, anchor=south, /pgf/number format/.cd, fixed, fixed zerofill, precision=1},
]
\addplot[fill=barOrangeDark, draw=white] coordinates {(Acc.\,A$\uparrow$,73.0) (Acc.\,B$\uparrow$,27.9) (Asym.$\downarrow$,44.2)};
\addlegendentry{D}
\addplot[fill=barOrangeLight, draw=white] coordinates {(Acc.\,A$\uparrow$,77.6) (Acc.\,B$\uparrow$,23.4) (Asym.$\downarrow$,56.3)};
\addlegendentry{S}
\addplot[fill=barOrangeLighter, draw=white] coordinates {(Acc.\,A$\uparrow$,85.8) (Acc.\,B$\uparrow$,51.9) (Asym.$\downarrow$,36.3)};
\addlegendentry{E}
\end{axis}
\end{tikzpicture}
\caption{Accuracy and asymmetry score.}
\label{fig:us_accuracy}
\end{subfigure}
\hfill
\begin{subfigure}[b]{0.48\textwidth}
\centering
\begin{tikzpicture}
\begin{axis}[
    ybar, bar width=16pt,
    width=\textwidth, height=4cm,
    ymin=0, ymax=180,
    ylabel={\small Time (s)},
    ylabel style={font=\small},
    xtick={1,2,3},
    xticklabels={D, S, E},
    xticklabel style={font=\small},
    yticklabel style={font=\scriptsize},
    ytick={0,40,80,120,160},
    yticklabels={0,40,80,{},160},
    extra y ticks={120},
    extra y tick labels={},
    extra y tick style={grid=major, grid style={dotted, gray!50}, tick style={black}},
    xmin=0.5, xmax=3.5,
    ymajorgrids=true, grid style={dotted, gray!50},
    axis x line*=box, axis y line*=box,
    xtick pos=left, ytick pos=left,
    nodes near coords, nodes near coords style={font=\scriptsize, anchor=south, /pgf/number format/.cd, fixed, fixed zerofill, precision=1},
]
\addplot[draw=white, fill=barOrangeDark,    bar shift=0pt] coordinates {(1,114.2)};
\addplot[draw=white, fill=barOrangeLight,   bar shift=0pt] coordinates {(2, 90.7)};
\addplot[draw=white, fill=barOrangeLighter, bar shift=0pt] coordinates {(3,116.9)};
\end{axis}
\end{tikzpicture}
\caption{Verification time per task.}
\label{fig:us_time}
\end{subfigure}
\caption{User-study results ($n{=}30$, within-subjects). (a) Support-only (S) does not improve gold-B accuracy over the document baseline (D), while polarity-aware (E) raises both gold-A and gold-B accuracy and lowers the asymmetry score. (b) E imposes a modest time cost over S, reflecting added scrutiny of contradicting evidence.}
\label{fig:us_results}
\vspace{-10pt}
\end{figure*}

\begin{table}[t]
\centering\small\setlength{\tabcolsep}{5pt}
\begin{tabular}{llrrrr}
\toprule
\rowcolor{headcol}
\textbf{Measure} & \textbf{Contrast} & \textbf{W} & \textbf{$p$} & \textbf{$p_{\text{Holm}}$} & \textbf{$d$} \\
\midrule
\multirow{3}{*}{Gold-A}
 & D vs.\ S & 138.0 & .340 & .681 & 0.18 \\
 & S vs.\ E &  90.0 & .086 & .431 & 0.34 \\
 & D vs.\ E &  95.0 & .024 & .142 & 0.45 \\
\midrule
\multirow{3}{*}{Gold-B}
 & D vs.\ S & 137.5 & .135 & .431 & $-$0.16 \\
 & S vs.\ E &  58.0 & \textbf{.0003} & \textbf{.0029} & 0.82 \\
 & D vs.\ E &  45.5 & \textbf{.0010} & \textbf{.0076} & 0.72 \\
\midrule
\multirow{3}{*}{Asym.}
 & D vs.\ S & 118.5 & .090 & .431 & 0.27 \\
 & S vs.\ E &  86.0 & \textbf{.0045} & \textbf{.031} & $-$0.40 \\
 & D vs.\ E & 167.0 & .412 & .681 & $-$0.15 \\
\bottomrule
\end{tabular}
\caption{Full statistical results for the nine pre-registered accuracy/asymmetry tests (Wilcoxon signed-rank, $n{=}30$, per-participant differences). $p_{\text{Holm}}$: Holm--Bonferroni-corrected $p$ across the nine-test family. \textbf{Bold}: significant after correction ($p_{\text{Holm}}<.05$). The primary contrast (Gold-B, $S$ vs.\ $E$) and the $E$-vs-$D$ contrasts survive correction, whereas all $D$-vs-$S$ contrasts do not.}
\label{tab:us_full_stats}
\vspace{-15pt}
\end{table}

\vspace{0.12cm}
\noindent\textbf{Accuracy and asymmetry (H2, H3).} Two patterns emerge (\autoref{fig:us_results}, left; full statistics in \autoref{tab:us_full_stats}). First, the \textbf{false-completeness bias (H2)}: $S$ yields numerically lower gold-B accuracy than $D$ ($23.4\%$ vs.\ $27.9\%$) and a higher asymmetry score ($56.3$ vs.\ $44.2$); however, neither $D$-vs-$S$ difference reaches significance (gold-B: $W{=}137.5$, $p{=}.135$, $d{=}0.16$; asymmetry: $W{=}118.5$, $p{=}.090$, $d{=}0.27$; both n.s.\ after correction), so the bias is directionally consistent but not statistically established at this sample size. Second, polarity-aware grounding raises both accuracies: $E$ lifts gold-B accuracy to $51.9\%$ ($+28.5$~pp over $S$, $+24.0$~pp over $D$; both significant after correction, \autoref{tab:us_full_stats}), gold-A accuracy is also higher under $E$ ($85.8\%$; $+8.2$~pp over $S$, $+12.8$~pp over $D$, n.s.\ after correction), and the asymmetry score drops to $36.3$, the lowest of the three. The primary contrast (\textbf{H3}: $S$ vs.\ $E$ on gold-B accuracy) yields $d{=}0.82$ (Wilcoxon signed-rank $W{=}58.0$, $n{=}30$; $p{=}.0003$, Holm-corrected $p{=}.0029$), meeting the pre-registered $d \geq 0.8$ threshold and remaining significant after Holm--Bonferroni correction. H4's asymmetry component is supported in direction and significant for the $S$-vs-$E$ contrast ($W{=}86.0$, $p{=}.0045$, Holm-corrected $p{=}.031$), but $E$ does not return the asymmetry score to ${\sim}0$, indicating polarity-aware grounding substantially attenuates but does not eliminate the bias.

\vspace{0.12cm}
\noindent\textbf{Verification time (H1, H4).} Mean per-task time was $114.2$\,s ($D$), $90.7$\,s ($S$), and $116.9$\,s ($E$) (\autoref{fig:us_results}, right). Time results provide \emph{partial} support for \textbf{H1}: $S$ reduces time vs.\ $D$ ($-23.5$\,s; $W{=}118$, $p{=}.018$) as predicted, but $E$ does not ($+2.7$\,s vs.\ $D$, n.s.; $+26.2$\,s vs.\ $S$, $p{=}.052$), counter to the $E$ component of H1 and the time component of \textbf{H4}. We attribute this to the short contexts used in the study: each task's cited documents are brief, so $E$'s localization advantage of jumping directly to the relevant evidence yields little time saving, while the added red highlights prompt participants to scrutinize conflicting evidence they would otherwise skip. The time difference thus reflects added scrutiny rather than inefficiency, and is precisely what underlies $E$'s large gold-B accuracy gain. In realistic long-context deployments, where users must scan lengthy documents to locate conflicting evidence, we expect $E$'s direct localization to recover and likely reverse this gap; we leave a long-context timing study to future work.

\vspace{0.12cm}
\noindent\textbf{Order and learning effects.} A mixed-effects model with condition as fixed effect and participant and task as random intercepts shows that adding trial order as a covariate does not materially change condition effect sizes ($\leq 0.5$~pp on every measure), confirming that the Williams Latin square absorbs order effects. The $E$~vs.\ $S$ gap on gold-B accuracy is stable across the six order sequences (range: $26.1$--$30.4$~pp).

\vspace{0.12cm}
\noindent\textbf{Scope of the $S$ vs.\ $E$ gain.} The $S \to E$ improvement combines two changes: color-encoded polarity (red vs.\ green) and the act of pointing to contradicting passages. We do not attempt to separately attribute the gain, as both are direct consequences of the polarity signal itself; isolating them would require interface variants withholding one while preserving the other, which we leave to follow-up work.

\vspace{0.12cm}
\noindent\textbf{Limitations of scale.} The study is powered for the primary $S{\to}E$ contrast ($d \geq 0.8$) but underpowered for smaller effects. The non-significant $D$-vs-$S$ contrasts (\autoref{tab:us_full_stats}) are consistent with a small true effect ($d{=}0.16$ on gold-B accuracy) that this sample is not designed to detect; confirming the false-completeness bias, as well as domain- or task-type-specific differences between conditions, would require substantially larger samples (\S\hyperref[sec:limitations]{Limitations}).

\vspace{0.12cm}
\noindent\textbf{Hypothesis outcomes.} We summarize the pre-registered hypotheses against observed results: \textbf{H1} (sentence-level grounding reduces time vs.\ $D$): \emph{partially supported}; $S$ reduces time as predicted, $E$ does not. \textbf{H2} ($S$ induces asymmetric bias vs.\ $D$): \emph{not statistically supported}; the differences are directionally consistent ($S$ shows lower gold-B accuracy and a higher asymmetry score than $D$) but do not reach significance ($p{=}.135$ and $p{=}.090$), consistent with a small effect under a sample powered for large effects (\S\hyperref[sec:limitations]{Limitations}). \textbf{H3, primary} ($E$ exceeds $S$ on gold-B accuracy with $d \geq 0.8$): \emph{supported}; $d{=}0.82$, significant after Holm--Bonferroni correction. \textbf{H4} ($E$ returns asymmetry to ${\sim}0$ without increasing time vs.\ $S$): \emph{partially supported}; $E$ significantly lowers the asymmetry score relative to $S$ ($p{=}.031$ after correction) but does not reach ${\sim}0$, and verification time increases rather than remains unchanged relative to $S$.

\clearpage
\onecolumn
\section{Instructions And Demonstration} \label{sec:appendix5}\label{sec:prompts}

\subsection{Prompt For Zero-Shot} \label{sec:appendix5.3}
\begin{center}
    \small
    \begin{tabular}{p{\textwidth}}
    \toprule
    \textbf{Task:} \\
    You are an expert fact-checking annotator. \\ \\

    You are given a RESPONSE consisting of several numbered sentences, and a CONTEXT consisting of several numbered sentences. \\ \\

    For each response sentence, identify which context sentences SUPPORT it and which context sentences CONTRADICT it. Ignore irrelevant context sentences (do not list them). \\ \\

    \textbf{Output format:} \\
    Output a JSON object with the following structure: \\
    \{ \\
    \quad "1": \{"support": [<list of context sentence indices>], "contradict": [<list of context sentence indices>]\}, \\
    \quad "2": \{"support": [...], "contradict": [...]\}, \\
    \quad ... \\
    \} \\ \\

    Keys are response sentence indices (1-based). Values are objects with: \\
    - "support": list of 1-based context sentence indices that support this response sentence \\
    - "contradict": list of 1-based context sentence indices that contradict this response sentence \\ \\

    If no context sentence supports or contradicts a given response sentence, use an empty list []. \\
    Output ONLY the JSON object, no explanation. \\

    \\
    \hdashline
    \\

    \textbf{Input Response:} \\
    \{Response\}

    \\
    \textbf{Input Context:} \\
    \{Context\}

    \\
    \textbf{Output:} \\

    \bottomrule
    \end{tabular}
    \captionof{table}{Prompt instructions used by GPT-4o and Qwen3-235B-Instruct for zero-shot end-to-end evidence grounding.}

\end{center}

\clearpage

\subsection{Prompt For Seven-Shot} \label{sec:appendix5.4}
\begin{center}
    \small
    \begin{tabular}{p{\textwidth}}
    \toprule
    \textbf{Task:} \\
    You are an expert fact-checking annotator. \\ \\

    You are given a RESPONSE consisting of several numbered sentences, and a CONTEXT consisting of several numbered sentences. \\ \\

    For each response sentence, identify which context sentences SUPPORT it and which context sentences CONTRADICT it. Ignore irrelevant context sentences (do not list them). \\ \\

    \textbf{Output format:} \\
    Output a JSON object with the following structure: \\
    \{ \\
    \quad "1": \{"support": [<list of context sentence indices>], "contradict": [<list of context sentence indices>]\}, \\
    \quad "2": \{"support": [...], "contradict": [...]\}, \\
    \quad ... \\
    \} \\ \\

    Keys are response sentence indices (1-based). Values are objects with: \\
    - "support": list of 1-based context sentence indices that support this response sentence \\
    - "contradict": list of 1-based context sentence indices that contradict this response sentence \\ \\

    If no context sentence supports or contradicts a given response sentence, use an empty list []. \\
    Output ONLY the JSON object, no explanation. \\ \\

    Here are seven examples across different domains: \\

    \\
    \hdashline
    \\

    \textbf{Example 1 (bioasq)} \\
    \textbf{RESPONSE:} \\
    1. CPX-351 (United States: Vyxeos\textregistered; Europe: Vyxeos\textregistered\ Liposomal), a dual-drug liposomal encapsulation of daunorubicin and cytarabine in a synergistic 1:5 molar ratio, is approved by the US FDA and the EMA for the treatment of adults with newly diagnosed therapy-related acute myeloid leukemia or acute myeloid leukemia with myelodysplasia-related changes. \\ \\

    \textbf{CONTEXT:} \\
    1. Quality-adjusted Time Without Symptoms of disease or Toxicity (Q-TWiST) analysis of CPX-351 versus 7 + 3 in older adults with newly diagnosed high-risk/secondary AML. \\ \\
    
    2. BACKGROUND: CPX-351 (United States: Vyxeos\textregistered; Europe: Vyxeos\textregistered\ Liposomal), a dual-drug liposomal encapsulation of daunorubicin and cytarabine in a synergistic 1:5 molar ratio, is approved by the US FDA and the EMA for the treatment of adults with newly diagnosed therapy-related acute myeloid leukemia or acute myeloid leukemia with myelodysplasia-related changes. \\ \\
    
    \textit{... (context sentences 3--12 omitted for space) ...} \\ \\
    
    13. CPX-351, marketed as Vyxeos, is a liposomal formulation that encapsulates daunorubicin and cytarabine at a fixed 5:1 molar ratio of daunorubicin to cytarabine. \\ \\
    
    14. It received FDA approval for the treatment of adults with newly diagnosed acute lymphoblastic leukemia (ALL) who carry the Philadelphia chromosome translocation, rather than for myeloid malignancies. \\ \\

    \textbf{OUTPUT:} \\
    \textcolor{myBlue}{\{"1": \{"support": [2], "contradict": [13, 14]\}\}} \\

    \\
    \hdashline
    \\

    \textit{(Examples 2--7, covering the fiqa, lifestyle, recreation, science, technology, and writing domains, follow the same format and are omitted here for space.)} \\

    \\
    \hdashline
    \\

    \textbf{Input Response:} \\
    \{Response\}

    \\
    \textbf{Input Context:} \\
    \{Context\}

    \\
    \textbf{Output:} \\

    \bottomrule
    \end{tabular}
    \captionof{table}{Prompt instructions and demonstrations used by GPT-4o and Qwen3-235B-Instruct for seven-shot end-to-end evidence grounding.}

\end{center}

\clearpage

\end{document}